\documentclass{article} 
\usepackage{iclr2025_conference,times}
\newcommand{\cgap}[1]{\raisebox{0.1em}{\scriptsize #1}\,}

\newcommand{\datanamefull}{multigranular textual descriptions}
\newcommand{\datasetnamefull}{MedTrinity-25M}
\newcommand{\llmnamefull}{multimodal large language model}
\newcommand{\llmname}{MLLMs}
\newcommand{\captioner}{LLaVA-Medcap}
\newcommand{\modelnamefull}{LLaVA-Tri}

\newcommand{\llavamed}{LLaVA-Med}
\newcommand{\llama}{LLaMA3}

\newcommand{\RN}[1]{%
	\textup{\lowercase\expandafter{\it \romannumeral#1}}%
}

\usepackage{amsmath,amsfonts,bm}

\def\eqref#1{equation~\ref{#1}}

\def\1{\bm{1}}

\DeclareMathAlphabet{\mathsfit}{\encodingdefault}{\sfdefault}{m}{sl}
\SetMathAlphabet{\mathsfit}{bold}{\encodingdefault}{\sfdefault}{bx}{n}

\usepackage{hyperref}       
\usepackage{url}            
\usepackage{booktabs}       
\usepackage{amsfonts}       
\usepackage{caption}
\usepackage{longtable}
\usepackage{subcaption}
\usepackage{graphicx}
\usepackage{multirow}
\usepackage{graphicx}
\usepackage{geometry}
\usepackage{amsmath}
\usepackage{cleveref}
\usepackage{float}
\crefname{figure}{Figure}{Figures}
\crefname{table}{Table}{Tables}
\crefname{section}{Section}{Sections}
\crefname{appendix}{Appendix}{Appendix}

\title{MedTrinity-25M: A Large-scale Multimodal Dataset with Multigranular Annotations for Medicine}
\usepackage{colortbl}
\usepackage{makecell}
\usepackage{pifont}

\author{%
\bf Yunfei Xie$^{1,*}$ \, Ce Zhou$^{1,\ast}$ \, Lang Gao$^{1,\ast}$ \, Juncheng Wu$^{2,\ast}$ \, Xianhang Li$^2$\\
\bf Hong-Yu Zhou$^3$ \, Sheng Liu$^4$ \, Lei Xing$^4$ \, James Zou$^4$ \, Cihang Xie$^2$ \, Yuyin Zhou$^2$ 
\vspace{.3em} \\
$^{\star}$equal technical contribution\vspace{.5em} \\
$^1$Huazhong University of Science and Technology  \quad\quad
$^2$UC Santa Cruz \\
$^3$Harvard University \quad\quad
$^4$Stanford University
}

\iclrfinalcopy 
\begin{document} 

\maketitle

\begin{abstract}
This paper introduces \datasetnamefull, a comprehensive, large-scale multimodal dataset for medicine, covering over 25 million images across 10 modalities with multigranular annotations for more than 65 diseases. These multigranular annotations encompass both global information, such as modality and organ detection, and local information like ROI analysis, lesion texture, and region-wise correlations. Unlike the existing multimodal datasets, which are limited by the availability of image-text pairs, we have developed the first automated pipeline that scales up multimodal data by generating multigranular visual and textual annotations in the form of image-ROI-description triplets without the need for any paired text descriptions. Specifically, data from over 30 different sources has been collected, preprocessed, and grounded using domain-specific expert models to identify ROIs related to abnormal regions. We then build a comprehensive knowledge base and prompt multimodal large language models to perform retrieval-augmented generation with the identified ROIs as guidance, resulting in multigranular textual descriptions. Compared with existing datasets, MedTrinity-25M provides the most enriched annotations, supporting a comprehensive range of multimodal tasks such as captioning and report generation, as well as vision-centric tasks like classification and segmentation. We propose \modelnamefull~by pretraining LLaVA on \datasetnamefull, achieving state-of-the-art performance on VQA-RAD, SLAKE, and PathVQA, surpassing representative SOTA multimodal large language models. Furthermore, \datasetnamefull~can also be utilized to support large-scale pre-training of multimodal medical AI models, contributing to the development of future foundation models in the medical domain. The dataset is publicly available at~\url{https://yunfeixie233.github.io/MedTrinity-25M/}.
\end{abstract}

\section{Introduction}
\label{sec:introduction}

Large-scale multimodal foundation models~\citep{visual-instruction,achiam2023gpt,tu2024towards,team2023gemini,zhou2024generalist} have demonstrated remarkable success across various domains due to their ability to understand complex visual patterns in conjunction with natural language. This success has sparked significant interest in applying such models to medical vision-language tasks.
Much progress has been made in improving the medical capacity of general domain multimodal foundation models by constructing medical datasets with image-text pairs and fine-tuning general domain models on these datasets~\citep{Bustos_2020-padchest, irvin2019chexpert, johnson2019mimiccxrjpg,li2024llava,quilt}.
However, current medical datasets have several limitations. Firstly, these datasets lack \textbf{multigranular} annotations that reveal the correlation between region-wise information within medical images. 
Medical images often contain detailed cues, such as regional abnormal textures or structures, which may indicate specific types of lesions. Therefore, multimodal models need the ability to infer global information, such as disease or lesion type, from local details. The absence of such data limits the models' capacity to comprehensively understand medical images.
%
Moreover, current dataset construction methods heavily rely on medical images paired with reports or captions from human experts~\citep{quilt, liu2021slake, lau2018datasetvqarad, he2020pathvqa}, which restricts their scalability. 
In this paper, we address the above challenges by proposing an automated data construction pipeline using \llmnamefull s~(\llmname) without relying on paired text descriptions.
To address the scarcity of medical knowledge within general-purpose MLLMs, we incorporate retrieval-augmented generation (RAG) to source relevant medical knowledge from a medical database for \llmname’s reference.
To enhance the model’s regional focus, we employ an ensemble of domain-specific segmentation models and grounding models to generate regions of interest (ROIs). MLLMs are then prompted to produce multigranular visual and textual annotations, enriched by the retrieved medical knowledge and ROIs. 
Our proposed pipeline enables the transformation of large-scale images without paired ROIs or text into image-ROI-description triplets. These triplets provide multigranular annotations that encompass both global textual information, such as disease/lesion type, modality, and inter-regional relationships, as well as detailed local annotations for ROIs, including bounding boxes, segmentation masks, and region-specific textual descriptions.
Using the proposed pipeline, we create a large-scale multimodal multigranular medical dataset containing over 25 million triplets, namely \textbf{\datasetnamefull}. To the best of our knowledge, this is the largest multimodal dataset in medicine to date.
 
To demonstrate the effectiveness of our dataset, we propose \textbf{\modelnamefull} by pretraining LLaVA on~\datasetnamefull. We conduct extensive evaluations across three external medical visual QA datasets representing different sub-pathologies. \modelnamefull~achieved state-of-the-art results in all three VQA benchmarks, with 75.0\% accuracy on VQA-RAD~\citep{lau2018datasetvqarad}, 87.8\% on SLAKE~\citep{liu2021slake}, and 65.3\% on PathVQA~\citep{he2020pathvqa}. Moreover, consistent performance improvements are observed when pretraining other multimodal models on~\datasetnamefull. These findings emphasize the potential of \datasetnamefull~as a foundational dataset that can improve the medical performance of diverse multimodal models.

\section{Related Work}
\label{sec:related_work}
\textbf{Medical Multimodal Foundation Models.} Due to the success of multimodal foundation models in comprehending visual features, their adaptation for medical vision-language tasks has garnered increasing attention~\citep{med-flamingo,tu2023generalist,li2024llava,zhou2024generalist}. Several works adapt general multimodal models to the medical domain via end-to-end training on medical datasets. For instance, Med-Flamingo~\citep{med-flamingo} fine-tunes OpenFlamingo-9B~\citep{Awadalla2023OpenFlamingoAO} using 0.8M interleaved and 1.6M paired medical image-text data. LLaVA-Med~\citep{li2024llava} uses a two-stage end-to-end visual instruction tuning~\citep{visual-instruction}, excelling in medical visual question answering (VQA) tasks. Med-Gemini~\citep{saab2024capabilities} adapts Gemini~\citep{gemini} using a long-form question-answer dataset to enhance multimodal and long-context capabilities. Despite these achievements, the limited scale of training data remains a challenge. Prior research~\citep{scalingLaws} shows that increasing training data size improves large multimodal model performance. This paper aims to build a large-scale medical dataset to drive the development of stronger medical multimodal foundation models.

\textbf{Multimodal Datasets for Medicine.}
The importance of constructing medical multimodal datasets has drawn significant attention~\citep{li2024llava,Pelka2018-ke-roco,zhang2024radgenome,irvin2019chexpert}. 
Several works focus on collecting images paired with clinical reports from specialists~\citep{zhang2024radgenome,irvin2019chexpert,johnson2019mimiccxrjpg}, providing detailed descriptions, including disease types and reasoning. 
For instance, MIMIC-CXR~\citep{johnson2019mimiccxrjpg} contains 227,835 images for 65,379 patients, with corresponding reports.
However, constructing such reports manually is time-consuming and costly, limiting dataset size. 
PMC-OA~\citep{lin2023pmc} includes up to 1.65 million image-caption pairs from medical papers but lacks detailed clinical reports. 
RadGenome-Chest CT~\citep{zhang2024radgenome} offers richer annotations but remains dependent on paired image-text data, limiting its scale. 
In comparison, we introduce the first automated pipeline to generate multigranular annotations for independent images, generating a comprehensive dataset containing 25 million samples.

\section{MedTrinity-25M Dataset}
\label{sec:method}

\subsection{Data Triplet}
\label{subsec:data_formatting}
In this section, we provide details about the data format within \datasetnamefull.
Our dataset comprises triplets of $\{\texttt{image}, \texttt{ROI}, \texttt{description}\}$. For each \texttt{image}, we provide multigranular annotations containing both textual \texttt{description} and visual \texttt{ROI}.

\begin{figure}[tb!]  
    \centering  
    \begin{subfigure}[b]{\textwidth}  
        \includegraphics[width=.9\textwidth]{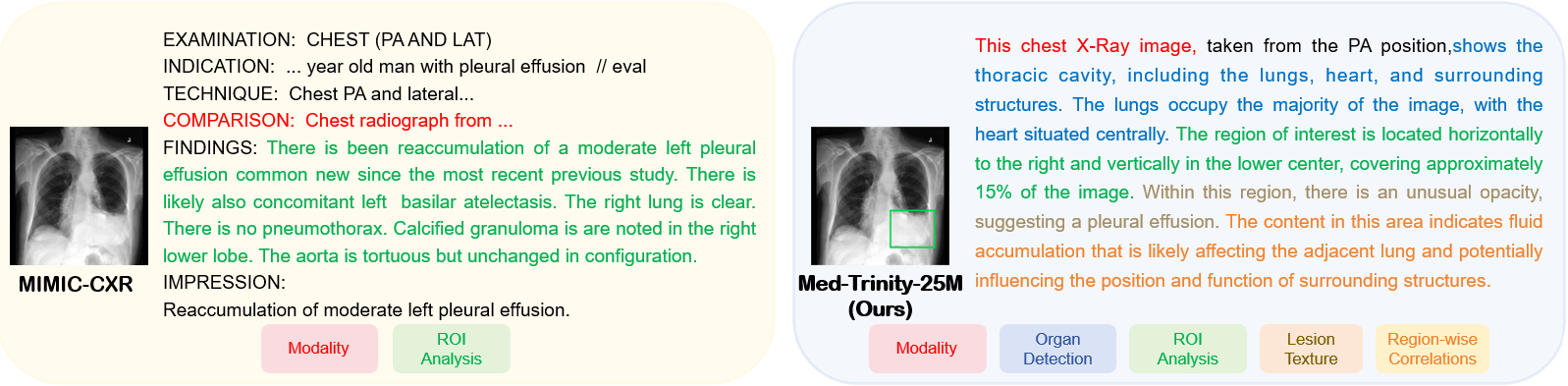}  
        \caption{Qualitative Comparison with sample in radiology report of chest x-rays dataset MIMIC-CXR~\citep{johnson2019mimic}.}  
        \label{fig:compare_mimic}  
    \end{subfigure}  
    \hspace{1em} 
    \begin{subfigure}[b]{\textwidth}  
        \includegraphics[width=.9\textwidth]{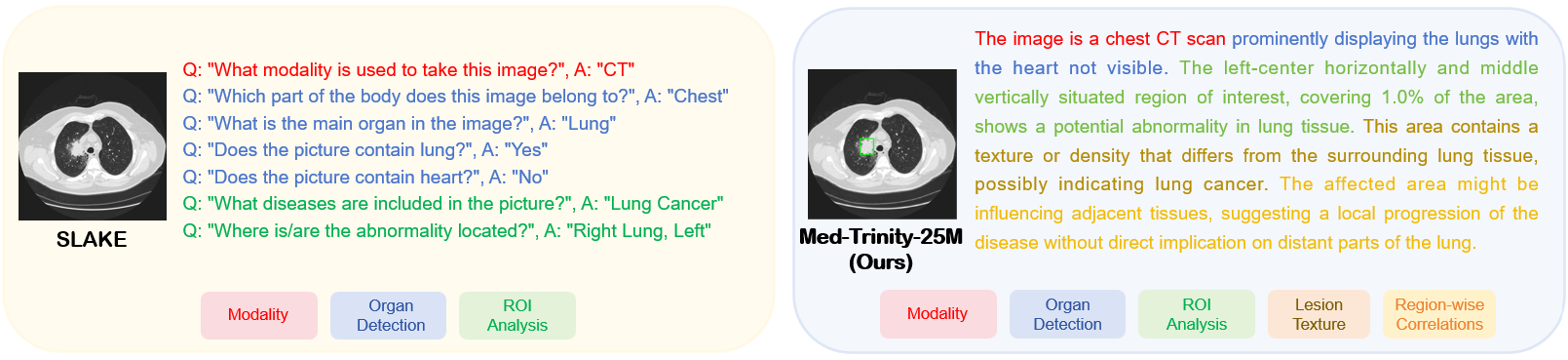}  
        \caption{Qualitative Comparison with sample in visual QA dataset SLAKE~\citep{liu2021slake}.}  
        \label{fig:compare_slake}  
    \end{subfigure}  
    \caption{Qualitative comparison with different types of dataset.}  
    \label{fig:data_compare}  
\end{figure}

\paragraph{Images.} 
We gather 25,016,845 samples across 10 medical image modalities and over 65 diseases. 
Specifically, we utilize original medical images from various datasets, extensively collecting from online sources such as \href{https://www.cancerimagingarchive.net/}{TCIA}, \href{https://www.kaggle.com/}{Kaggle}, \href{https://zenodo.org/}{Zenodo}, \href{https://www.synapse.org/}{Synapse}, \href{https://huggingface.co/}{Hugging Face}, \href{https://grand-challenge.org/}{Grand Challenge}, \href{https://github.com/}{GitHub}, and medical datasets, including CheXpert~\citep{irvin2019chexpert} and DeepLesion~\citep{deeplesion}. 
3D volumetric images in DICOM or NIfTI formats are converted to 2D slices in PNG format.
The detailed data sources are illustrated in ~\cref{sec:datasource}.
\paragraph{ROIs.}
We use ROIs to provide visual annotations for each image, primarily focusing on pathological findings such as lesions, inflammation, neoplasms, infections, and other abnormalities. 
In cases without such abnormalities, the ROIs generally mark the primary object or organ in the image, as illustrated in ~\cref{fig:normal_roi_combined}.
When multiple organs are relevant for disease diagnosis, the ROIs aim to cover several regions associated with the disease, providing detailed analysis of each affected area, as shown in ~\cref{fig:global_local_example}.
\paragraph{Textual Descriptions.}
The textual descriptions for each image are composed of detailed information across various attributes. 
In contrast to the unstructured medical reports or short captions in previous medical datasets \citep{irvin2019chexpert,johnson2019mimiccxrjpg,Bustos_2020-padchest,zhang2023pmcvqa,liu2021slake,Pelka2018-ke-roco,noauthor_undated-bj-pmc-oa}, our textual descriptions are structured and contain multigranular information for five attributes.
As illustrated in~\cref{fig:data_compare}, the general attributes of the image are described initially, covering aspects such as modality, the detection of specific organs, and their depiction. Following this, the attributes related to ROI are detailed, including the ROI analysis, locations and texture of the lesions, which encompass the disease type and relevant pathological features. Furthermore, region-wise correlations are highlighted to showcase relationships between the ROIs and surrounding regions, providing insight into differences in features and the extent of disease progression.

\subsection{Data Construction Pipeline}
\label{sec:pipeline}
\begin{figure*}[t]
  \centering
  \includegraphics[page=1, width=0.8\textwidth]{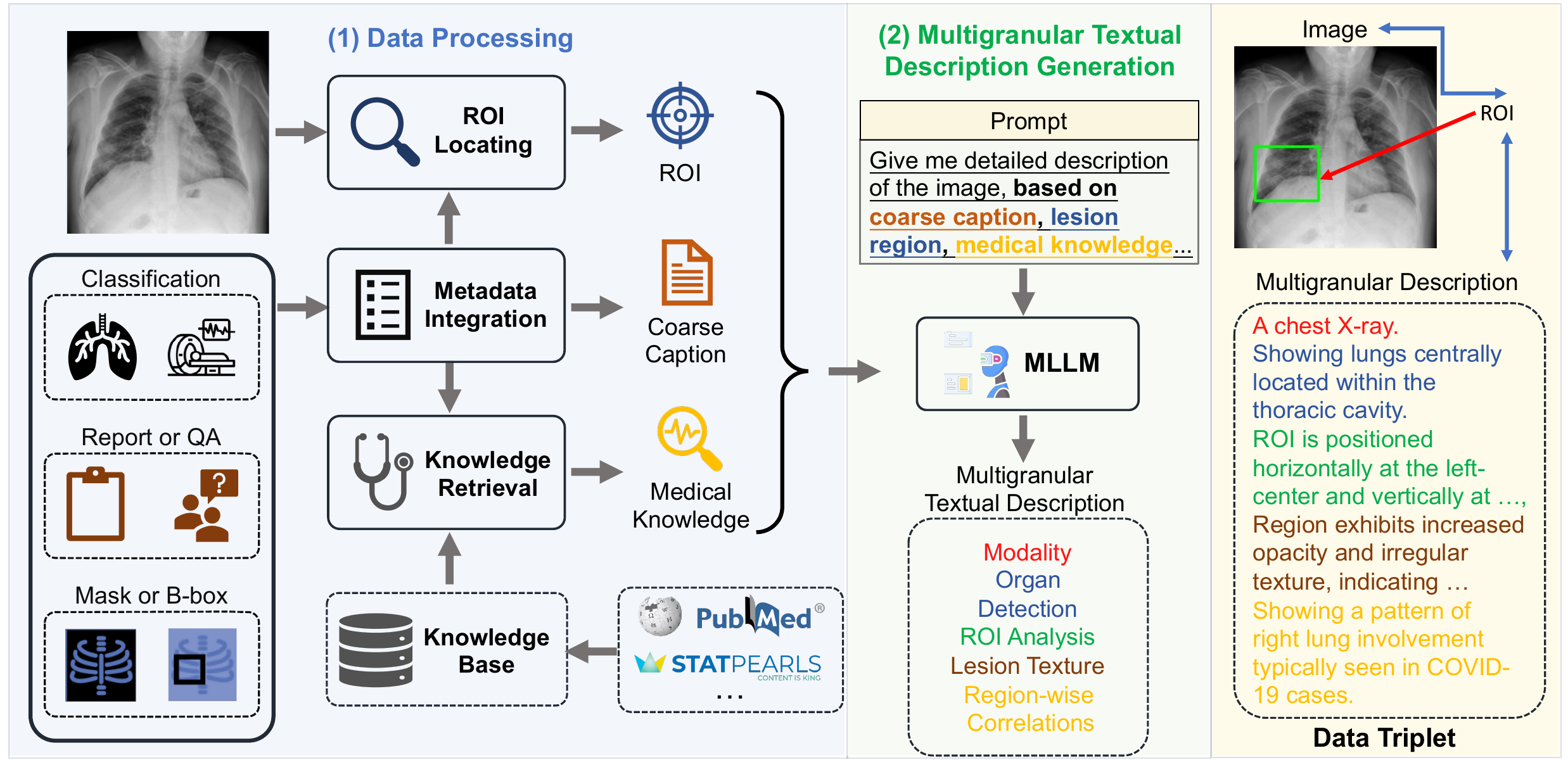}
  \caption{
  \textbf{Data construction pipeline.} \textit{1) Data processing}, including metadata integration to generate coarse caption, ROI locating, and medical knowledge collection. \textit{2) Multigranular Textual Description Generation} based on processed data.
  }
  \label{fig:pipeline}
  \vspace*{-0.5cm}
\end{figure*}

Given a medical image, we aim to generate corresponding multigranular visual and textual annotations. Specifically, as shown in \cref{fig:pipeline}, our pipeline can be decomposed into two stages: \textit{1) Data Processing}, and \textit{2) Multigranular Textual Description Generation}. Firstly, our data processing stage includes three key steps: \textit{a) Metadata Integration} to produce coarse captions encapsulating fundamental image information such as modality and disease types; \textit{b) ROI Locating} to identify regions of abnormalities; and \textit{c) Medical Knowledge Retrieval} to extract relevant fine-grained medical details. All processing steps are further detailed in~\cref{sec:data_preprocessing}.
Subsequently, we prompt \llmname~to integrate information within processed data and generate multigranular textual descriptions. Corresponding details are provided in \cref{sec:data_generation}.
The original image, generated visual ROIs, and textual descriptions are combined into a data triplet in \datasetnamefull.

\subsubsection{Data Processing}
\label{sec:data_preprocessing}

\paragraph{Coarse Caption Generation via Metadata Integration.} We aim to generate coarse captions that provide fundamental information for a given image, including modality, organ labels, disease types, and optionally, camera views and equipment information. Instead of extracting features directly from the images, we generate these captions by integrating dataset metadata. We first extract metadata from the datasets and then apply a fixed rule to integrate this information into coarse captions. For example, for an image in the QaTa-COV19 dataset\footnote{\url{https://www.kaggle.com/aysendegerli/qatacov19-dataset.}}, we derive metadata from the dataset's accompanying paper or documentation, indicating that it consists of COVID-19 chest X-ray images. Next, we construct coarse captions like ``A chest X-ray image with COVID-19 in the lungs'' highlighting the modality, organ types, and disease labels. We also integrate additional paired textual information (if any), such as radiological findings into coarse captions.

\begin{figure}[!t]
  \centering
  \includegraphics[page=1, width=0.71\textwidth]{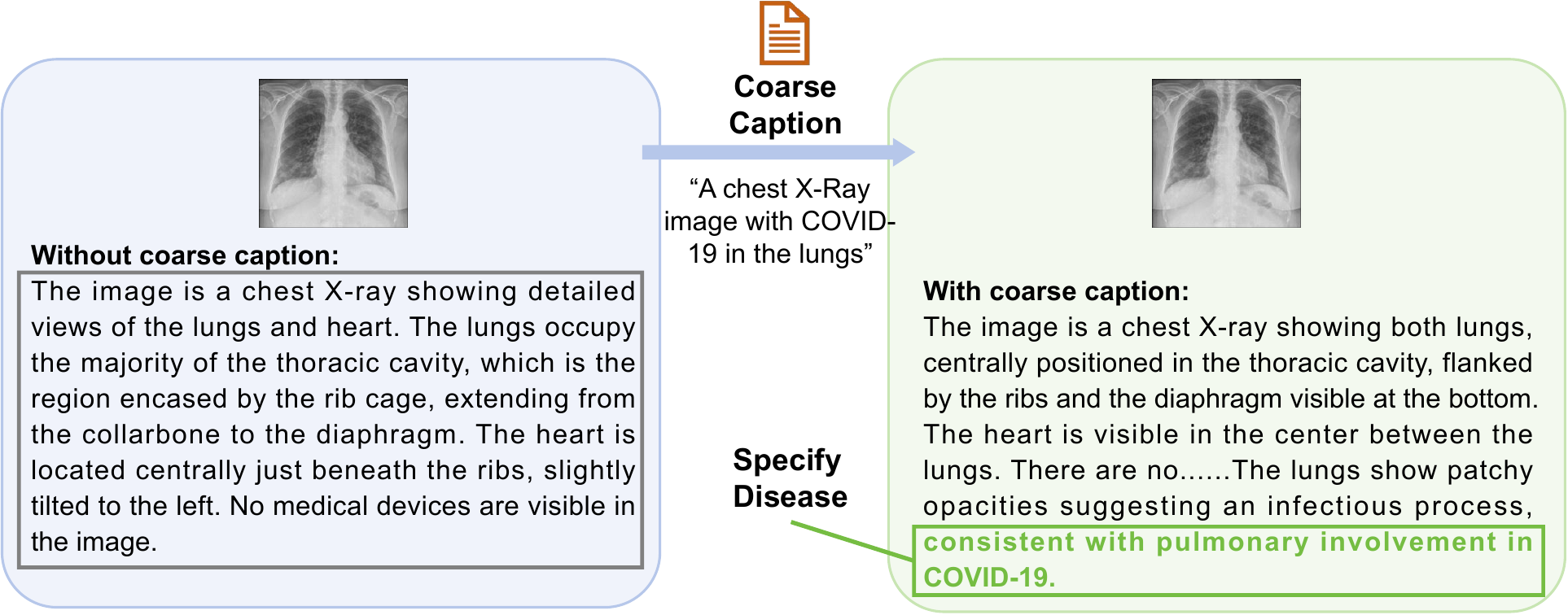}
  \caption{
  \textbf{A qualitative comparison example of generated textual description with and without coarse caption.} Without a coarse caption, \llmname~fails to detect diseases. On the contrary, providing a caption mentioning ``COVID-19'' allows \llmname~to identify and categorize the disease, facilitating further analysis. 
  }
  \label{fig:compare_coarse}
\end{figure}

The effectiveness of applying coarse captions when generating \datanamefull~is illustrated in~\cref{fig:compare_coarse}.
In contrast to the scenario without a coarse caption, where \llmname~fails to recognize the disease, providing \llmname~with a coarse caption that includes the disease type ``COVID-19'' enables it to identify and categorize the disease, thereby laying the foundation for further analysis.

\begin{figure}[!t]
  \centering
  \includegraphics[page=1, width=0.71\textwidth]{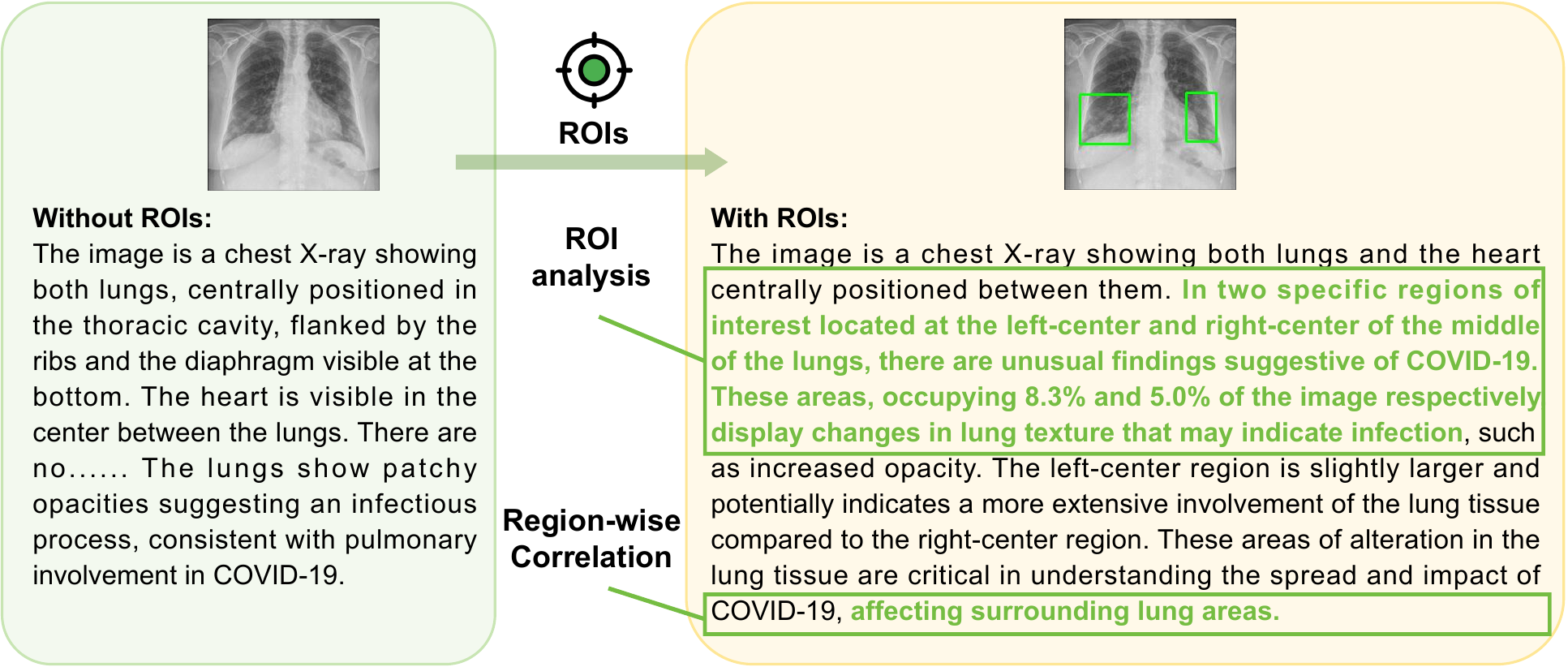}
  \caption{
  \textbf{A qualitative comparison example of generated textual description with and without locating ROIs.}  Without ROIs, the caption offers only a brief global analysis; with ROIs, \llmname~conducts detailed local analysis and assesses the impact of lesion ROIs on adjacent normal regions.
  }
  \label{fig:compare_roi}
  \vspace*{-0.5cm}
\end{figure}
\begin{figure}[!t]
  \centering
  \includegraphics[page=1, width=0.71\textwidth]{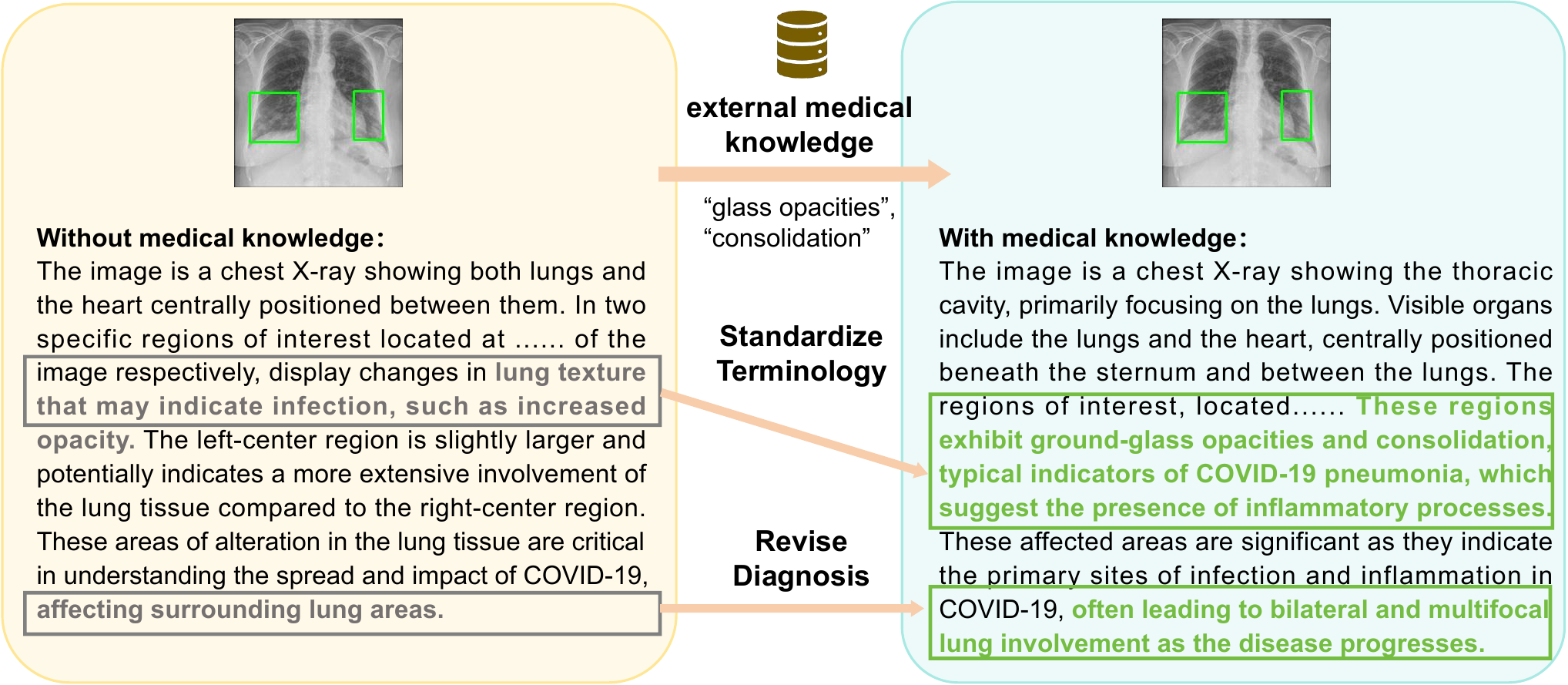}
  \caption{
  \textbf{A qualitative comparison example of generated textual description with and without external medical knowledge.} 
  \llmname~can standardize medical terminology in its expressions and refine its diagnosis based on disease progressions detailed in medical literature.}
  \label{-20pt}
  \label{fig:compare_rag}
  \vspace{-0.8cm}
\end{figure}

\paragraph{ROI Locating.}
\label{para:rois}

\begin{figure}[t]
    \centering
    \vspace*{0.5cm}
    \begin{subfigure}[b]{0.3\textwidth}
        \includegraphics[width=\textwidth]{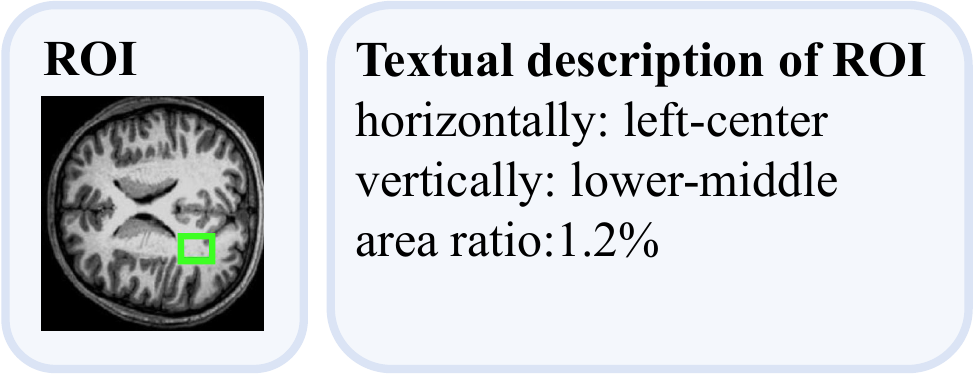}
        \caption{Example of locating ROI via SAT~\citep{zhao2023one}.}
        \label{fig:sat}
    \end{subfigure}
    \hspace{1em} 
    \begin{subfigure}[b]{0.3\textwidth}
        \includegraphics[width=\textwidth]{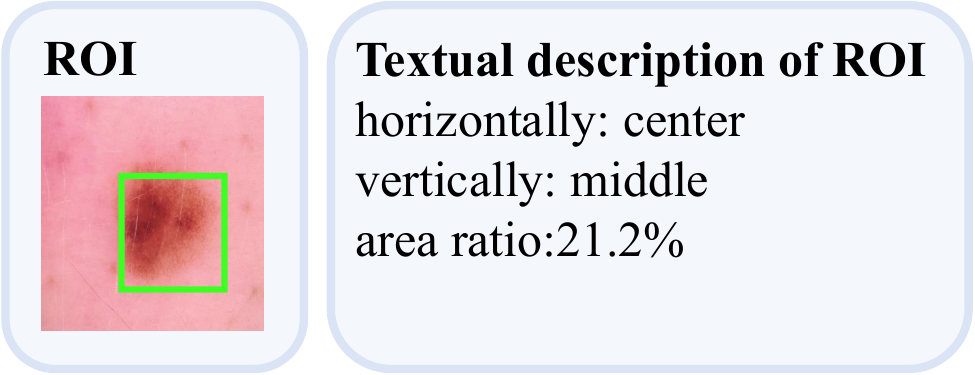}
        \caption{Example of locating ROI via BA-Transformer~\citep{Wang_2021_batransformer}.}
        \label{fig:specialized}
    \end{subfigure}
    \hspace{1em} 
    \begin{subfigure}[b]{0.3\textwidth}
        \includegraphics[width=\textwidth]{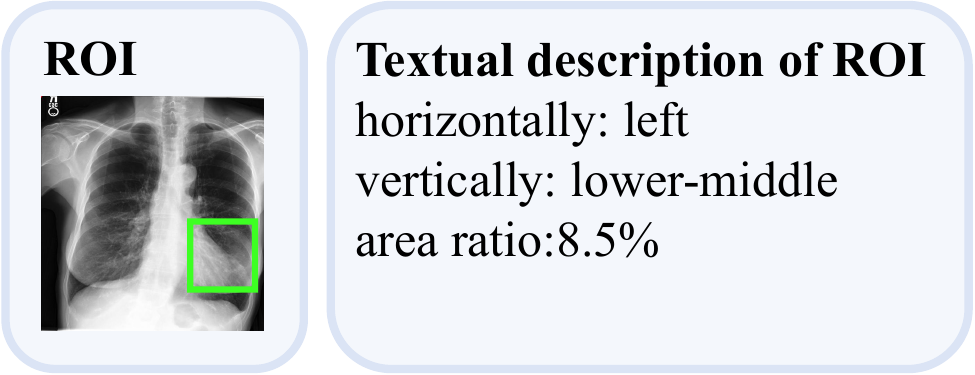}
        \caption{Example of locating ROI via Chexmask~\citep{Gaggion_2022_chexmask2}.}
        \label{fig:sam}
    \end{subfigure}
     \caption{Example of ROIs and their corresponding textual descriptions.}
    \label{fig:roi}
    \vspace*{-0.5cm}
\end{figure}

We employ appropriate strategies to locate ROIs for images paired with different annotations. 
For datasets that already include localization annotations, such as segmentation masks or bounding boxes, we derive the ROIs from these paired annotations. Specifically, bounding boxes are directly used as the ROIs, while segmentation masks are converted to ROIs by creating the smallest bounding box that covers the mask. 
When such localization annotations are not available, we apply corresponding pretrained expert models to generate ROIs. More details about the selection of expert models are provided in~\cref{sec:model_roi}.
Examples of generated ROIs from various modalities using corresponding models are demonstrated in~\cref{fig:roi}.  
For modalities such as X-ray and MRI scans viewed from the z-axis, our ROI localization employs a coordinate system relative to the human body, resulting in a left-right reversal in the image representation.

Incorporating ROIs as the guidance facilitates \llmname~to conduct a detailed analysis and generate \datanamefull.
As demonstrated in~\cref{fig:compare_roi}, description generated without guidance of ROIs is limited to a brief global overview of the image.
In comparison, with ROIs, generated description contains local analysis regarding the abnormal region and its correlations to other regions.

\begin{figure*}[t]
  \centering
  \includegraphics[page=1, width=.9\textwidth]{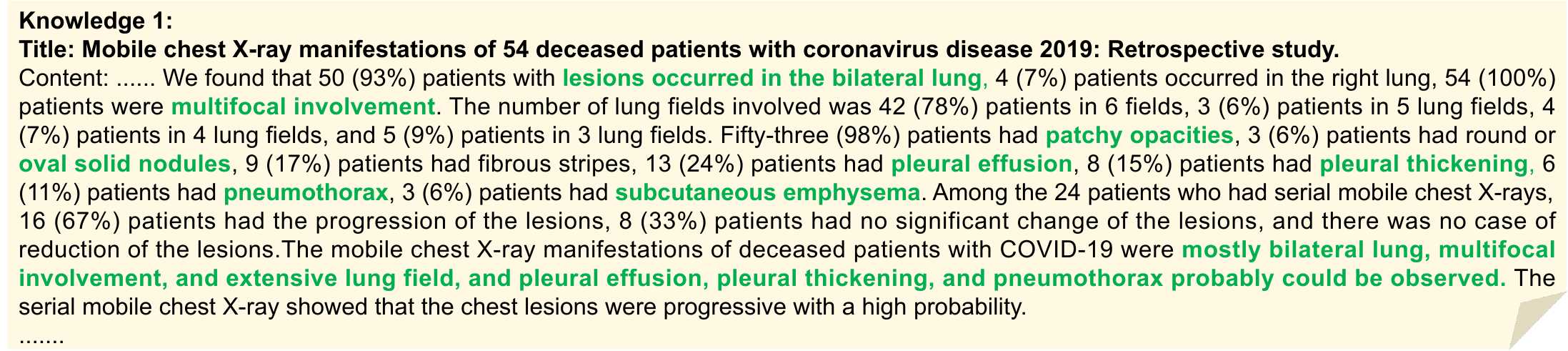}
  \caption{\textbf{An example of the Top-8 retrieval results.} By leveraging COVID-19-related medical knowledge, \llmname~can standardize medical terminology and enhance diagnoses according to the disease progressions described in medical literature.}
    \label{fig:knowledge_example}
    \vspace{-0.4cm}
\end{figure*}

\paragraph{Medical Knowledge Retrieval.}
General-purpose MLLMs often lack medical terminology and expertise.
To address this issue, we build a medical knowledge database following MedRAG~\citep{xiong2024benchmarking}. We collect three main corpora: PubMed\footnote{\url{https://pubmed.ncbi.nlm.nih.gov/}} for biomedical knowledge, StatPearls\footnote{\url{https://www.statpearls.com/}} for clinical decision support, and medical textbooks~\citep{jin2021disease} for domain-specific knowledge.
We segment these corpora into short snippets and encode them into high-dimensional vectors using the text encoder from Med-CPT~\citep{jin2023medcpt}. These vectors are then indexed into a specialized vector knowledge base using Faiss\citep{johnson2019billion}, optimized for efficient retrieval.
For a given image, we retrieve relevant medical knowledge by using its coarse caption, which is generated through metadata integration. Specifically, we encode the coarse captions, including disease and organ classifications, into vectors using the Med-CPT text encoder. We then perform a vector similarity search in the medical vector database, retrieving the top eight medical knowledge snippets that semantically match the query. These snippets provide the external medical knowledge paired with the image for generating textual descriptions. A qualitative example demonstrating the effectiveness of incorporating external medical knowledge is shown in \cref{fig:knowledge_example}. With access to COVID-19-related medical knowledge, \llmname~can standardize medical terminology and refine diagnoses based on the disease progressions outlined in medical literature.

A qualitative comparison of generated text descriptions, both with and without external medical knowledge, is presented in \cref{fig:compare_rag}. MLLMs are capable of standardizing medical terminology and enhancing diagnostic accuracy by incorporating insights from disease progressions documented in medical literature.

\subsubsection{Generation of Multigranular Text Description}
\label{sec:data_generation}

\paragraph{Generation Prompt.} After data processing, a comprehensive prompt is utilized to guide \llmname~to integrate all information and generate multi-granular descriptions.
We incorporate the processed data (coarse captions, ROIs, and retrieved medical knowledge) into the prompts.
Specifically, textual information such as coarse captions and retrieved medical knowledge are directly integrated into the prompt. 
While ROIs on images are converted into textual information based on their coordinates and area ratio within the images, using terms such as ``left-center'' and ``area ratio: 1.2\%''. Examples of textual information converted from ROIs are shown in \cref{fig:roi}.
Instead of merely inserting retrieved knowledge, we instruct MLLM to identify and align the relevant knowledge with ROIs to provide diagnostic insights.
The prompt template consists of a three-level hierarchical framework with questions to instruct \llmname~ to generate:
(1) a global description that captures all details of the image, (2) a local-focused analysis of specific ROIs that potentially are diseased;
and (3) an inference of the correlations between region-wise attributes to understand the impact of local abnormalities on the surrounding regions and extent of disease progression.
Detailed prompt template is presented in~\cref{sec:prompt}.

\paragraph{Choice of MLLM.}
All textual description in \datasetnamefull~are generated using \captioner, which is a specifically fine-tuned LLaVA to generate high-quality textual descriptions.
To obtain the fine-tuning data, we first generate 200,000 multigranular textual descriptions using our generation pipeline and GPT-4V~\citep{achiam2023gpt}.
Subsequently, we pretrain our \captioner~following the two-stage fine-tuning strategy in LLaVA-Med~\citep{li2024llava}, then these generated data are used to to finetune the \captioner.
\captioner~is then used in our pipeline to generate text descriptions for whole 25 million images in \datasetnamefull.
As shown in~\cref{sec:compare_gpt}, \captioner~is capable of generating high-quality descriptions with more details compared to GPT-4V.

\subsection{Dataset Analysis}

\begin{figure*}[t]
  \centering
  \begin{subfigure}[b]{0.45\textwidth}
    \centering
    \includegraphics[height=3.7cm]{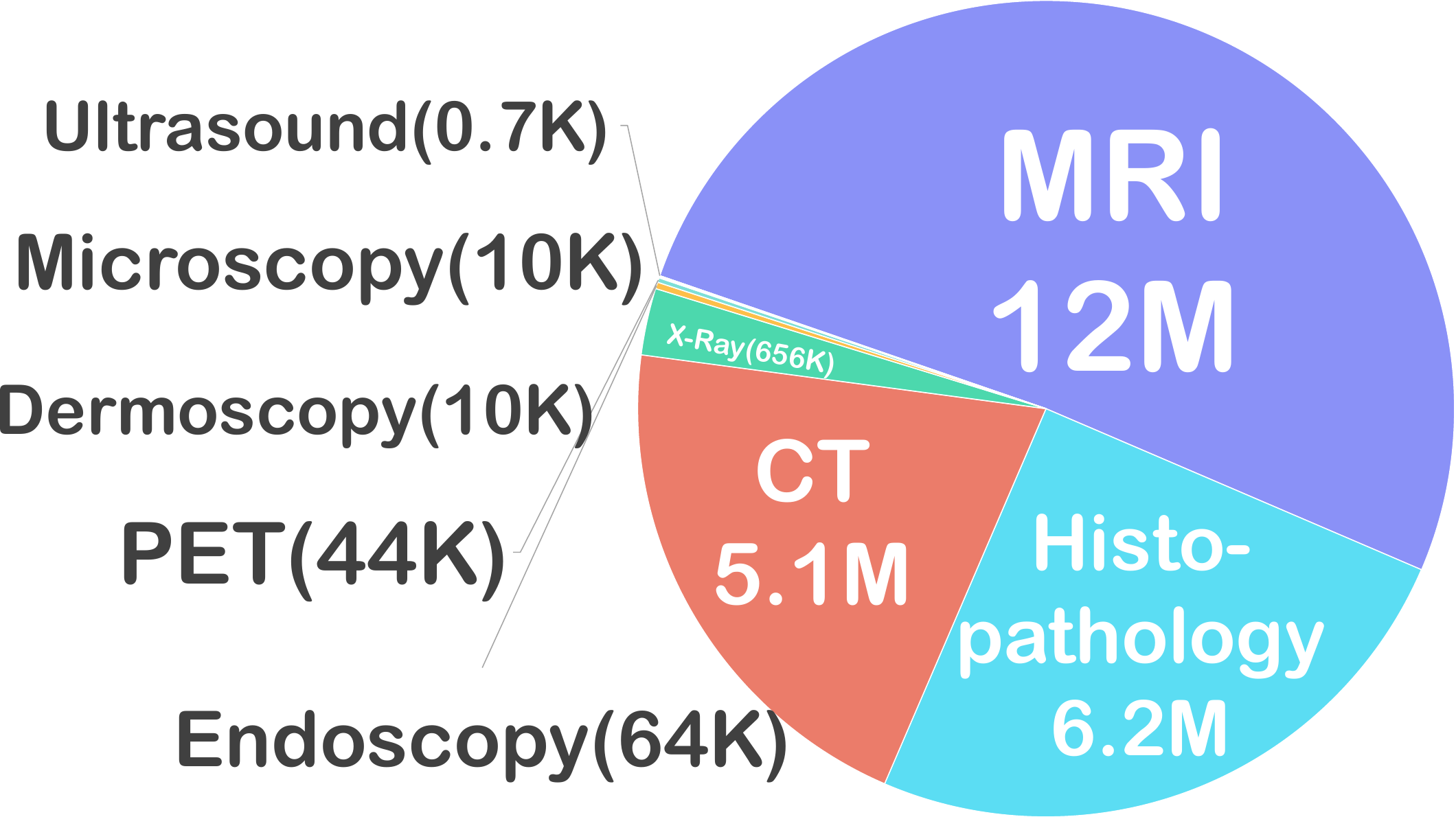}
    \caption{Modality distribution.}
    \label{fig:modalities}
  \end{subfigure} 
  \begin{subfigure}[b]{0.45\textwidth}
    \centering 
    \includegraphics[height=3.7cm]{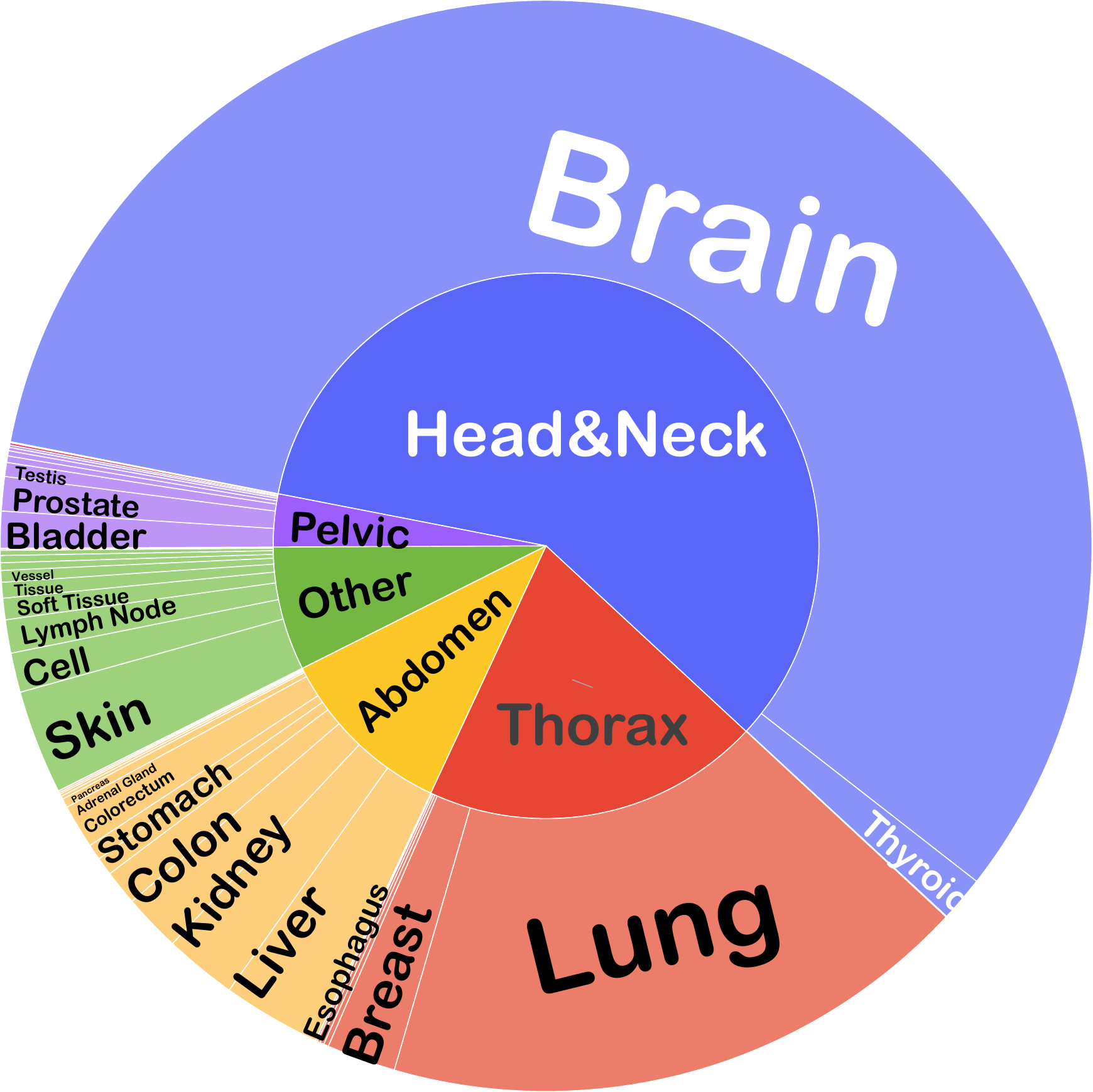}
    \caption{Biological structures.}
    \label{fig:distribution}
  \end{subfigure}
  
  \begin{subfigure}[b]{0.45\textwidth}
    \centering 
    \includegraphics[width=5cm]{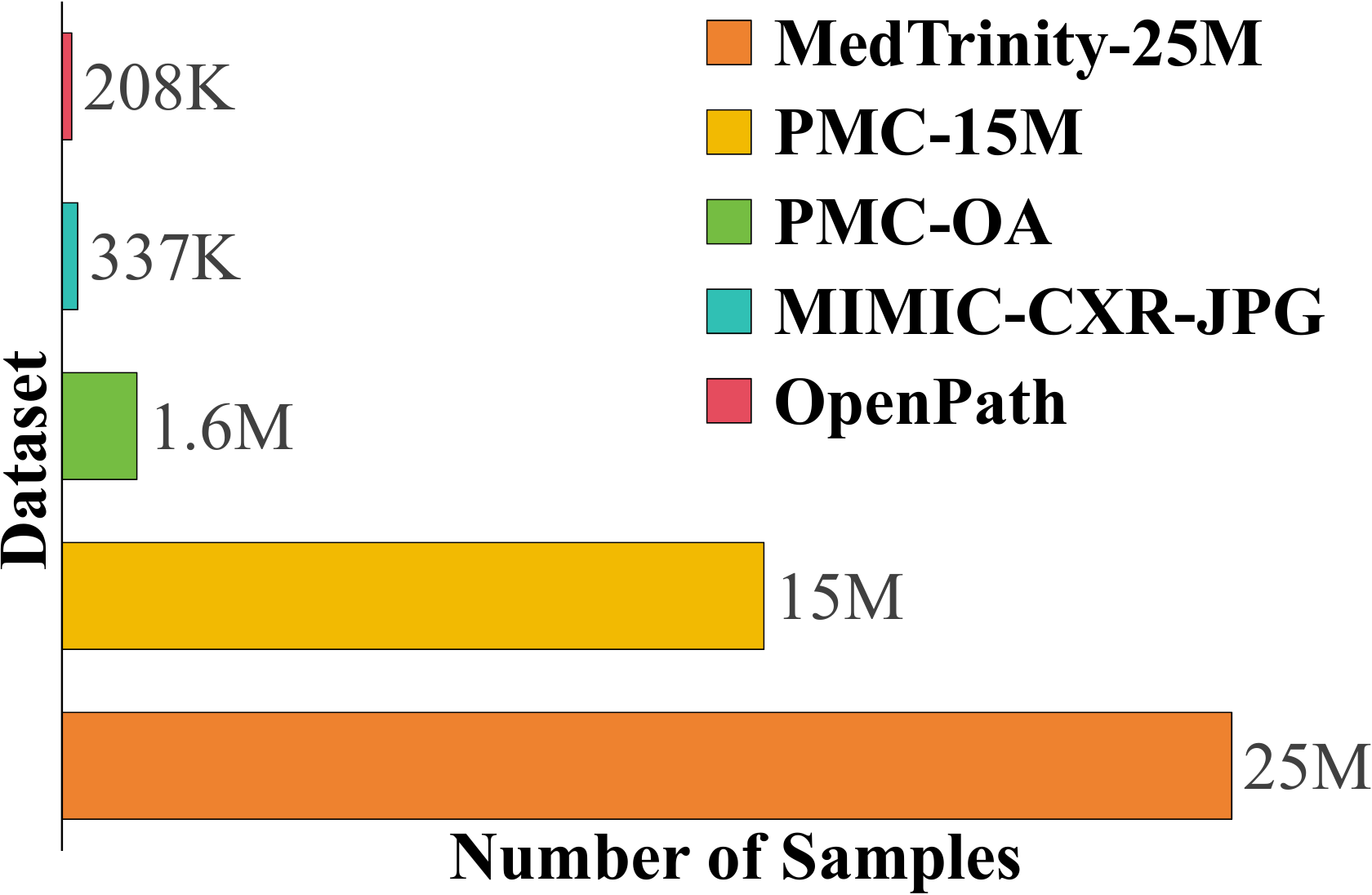}
    \caption{Data size comparison.}
    \label{fig:size}
  \end{subfigure}
   \begin{subfigure}[b]{0.45\textwidth}
    \centering
    \includegraphics[width=.9\textwidth]{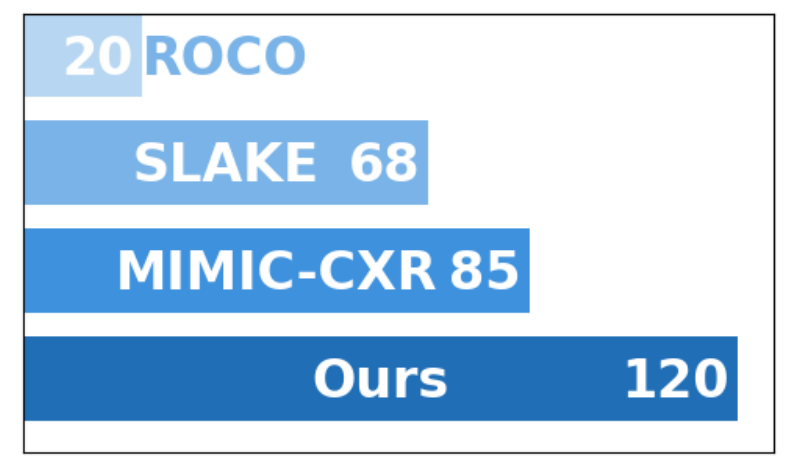}
    \caption{Average word count of descriptions comparison.}
    \label{fig:length}
  \end{subfigure}
   
  \begin{subfigure}[b]{\textwidth} 
    \centering
    \includegraphics[width=\textwidth]{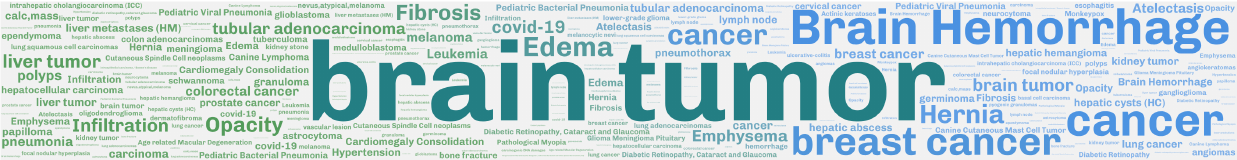}
    \caption{Wordcloud of disease statistic.}
    \label{fig:wordcloud}
\end{subfigure}
  \caption{Statistical overview of \datasetnamefull.}
  \label{fig:statistic}
  \vspace{-15pt}
\end{figure*}

\paragraph{Scale.} \cref{fig:size} compares the amount of data samples in \datasetnamefull~ and other medical multimodal datasets. To the best of our knowledge, \datasetnamefull~is the largest open-source, multi-modal multigranular medical dataset to date.

\paragraph{Diversity.} Our dataset encompasses 10 imaging modalties,  with more than 65 diseases across various anatomical structures in human. 
The number of the samples within each modality in \datasetnamefull~are shown in~\cref{fig:modalities}, and the distribution of each
anatomical and biological structures is provided in~\cref{fig:distribution}. 
Meanwhile, \cref{fig:wordcloud}  illustrates the frequently used words related to diseases in our dataset.

\paragraph{Richness.}
We provide both qualitative examples and quantitative analysis to demonstrate the richness of annotations in \datasetnamefull.
As shown in \cref{tab:compare_tick}, we compare the types of annotations in our dataset with those of other multimodal datasets. Our dataset provides multigranular and richer annotation information, surpassing other multimodal datasets. Qualitative examples are shown in \cref{fig:data_compare}. Our textual descriptions provide more comprehensive information compared to the chest X-rays dataset MIMIC-CXR~\citep{johnson2019mimic} and the visual QA dataset SLAKE~\citep{liu2021slake}.
\cref{fig:length} compares the average word count of text descriptions in multiple medical multimodal datasets. The word count in our dataset is significantly larger, indicating greater richness.

\begin{figure}[t]
        \captionof{table}{Comparison of types of annotations in \datasetnamefull~with other multimodal datasets.}  
    \centering

    \setlength{\tabcolsep}{2mm}
    \scriptsize 
    \begin{tabular}{ccccccc}
    \toprule
    Dataset & Modality  & \makecell[c]{Lesion\\ Type} & \makecell[c]{Lesion \\ BBox/Mask}   & \makecell[c]{Lesion \\ Description} & \makecell[c]{Region-wise \\  Correlations }\\
    \midrule
    MedMNIST~\citep{yang2023medmnist} & \ding{55} &\checkmark & \ding{55} & \ding{55} & \ding{55}\\
    DeepLesion~\citep{yan2017deeplesion} & \checkmark & \ding{55} & \checkmark  & \ding{55} & \ding{55}\\
    BraTS 2024~\citep{de20242024} & \checkmark & \ding{55} &\checkmark & \ding{55} & \ding{55}\\
    MIMIC-CXR~\citep{johnson2019mimic} & \checkmark &\checkmark &\checkmark &\checkmark & \ding{55}\\
    Quilt-1M~\citep{quilt} & \checkmark &\checkmark & \ding{55} &\checkmark & \checkmark\\
    VQA-RAD~\citep{lau2018dataset} & \checkmark & \checkmark & \ding{55} & \checkmark & \ding{55}\\
    CRC100K~\citep{Kather2018-aj-crc100k} & \checkmark & \checkmark & \ding{55} & \ding{55} & \ding{55}\\
    SA-Med2D-20M~\citep{sam20m} & \checkmark & \checkmark & \checkmark & \ding{55} & \ding{55}\\
    \rowcolor{lightgray}\textbf{\datasetnamefull (Ours)} & \checkmark & \checkmark & \checkmark & \checkmark & \checkmark\\
    \bottomrule
    \end{tabular}

    \label{tab:compare_tick} 
\vspace{-0.5cm}
\end{figure}%

\paragraph{Quality.}
We conduct expert and LLM evaluations to verify the quality of the generated multigranular descriptions. Each description in \datasetnamefull~contains five key attributes of medical images: modality, organ detection, ROI analysis, lesion texture, and region-wise correlations. A random subset of 200 samples is selected for evaluation. In expert-based evaluation, medical professionals assess the accuracy of each attribute by comparing the generated descriptions with ground-truth annotations. Scores are averaged across all samples to obtain an overall score. For LLM-based evaluation, we use GPT-4V to assess the accuracy of medical facts and diagnoses based on the same five attributes. GPT-4V scores each attribute on a scale of 0 to 2 points. All scores are normalized to a 0–1 scale for comparison.

\cref{tab:alignment} shows that \datasetnamefull~achieves 0.85 and 0.86 in expert and LLM evaluations, with modality, organ detection, and ROI analysis nearing perfect scores. To illustrate, \cref{fig:alignment_example} shows a sample that achieved a perfect score from GPT-4V.

\setlength{\belowcaptionskip}{5pt} 
\setlength{\dbltextfloatsep}{0pt}
\begin{table*}[!t]

\setlength{\tabcolsep}{1.5mm} 
\renewcommand{\arraystretch}{1} 
\scriptsize
\centering
\caption{Comparison of alignment scores between LLM and Expert.}
\resizebox{0.8\textwidth}{!}{
\begin{tabular}{lcccccc}
\toprule
\multirow{2}{*}{Evaluator} &
\multicolumn{6}{c}{Attributes} \\
\cline{2-7}
&
\makecell[c]{Modality} &
\makecell[c]{Organ \\ Detection} &
\makecell[c]{ROI \\ Analysis} &
\makecell[c]{Lesion \\ Texture} &
\makecell[c]{Region-wise \\ Correlations} &
Avg. \\
\midrule
LLM    & 1.00/1.00 & 0.90/1.00 & 0.90/1.00 & 0.80/1.00 & 0.70/1.00 & 0.86/1.00 \\
Expert & 1.00/1.00 & 0.90/1.00 & 0.90/1.00 & 0.70/1.00 & 0.80/1.00 & 0.85/1.00 \\
\bottomrule
\end{tabular}
}
\label{tab:alignment}
\vspace{-0.7cm}
\end{table*}
\section{Experiment}\label{sec:experiment}

\subsection{\modelnamefull: Aligning Multiscale MLLM with \datasetnamefull}
To fully exploit the multigranular annotations, we propose \modelnamefull, which is based on LLaVA~\citep{visual-instruction} and incorporates \datasetnamefull~to align it into the medical domain.
\modelnamefull~integrates \llama~\citep{llama3} to enhance linguistic capabilities and incorporates multiscale feature extraction \citep{shi2024we} to boost visual performance.
Specifically, we first pretrain \modelnamefull~on 600K image-text pairs from PMC-15M~\citep{zhang2023large}, following the training settings from~\citet{li2024llava}.
The model is then trained on \datasetnamefull 
 for multigranular alignment.

We benchmark \modelnamefull~on three biomedical Visual Question Answering (VQA) datasets, VQA-RAD~\citep{lau2018dataset}, SLAKE~\citep{liu2021slake}, and PathVQA~\citep{he2020pathvqa}, to assess the efficacy of aligning the model using \datasetnamefull.
Following LLaVA-Med, The model is fine-tuned for 15 epochs on each of the three VQA datasets and evaluated accordingly. As shown in \cref{tab:model_perf}, \modelnamefull~achieved state-of-the-art results in all of the three VQA benchmarks, with 75.0\% accuracy on VQA-RAD, 87.8\% on SLAKE, and 65.3\% on PathVQA. These results highlight the significant advantages of incorporating multiscale \modelnamefull~with multigranular alignment.

\subsection{Enhancing Model Performance through Multigranular Alignment}
To further demonstrate the effectiveness of multigranular alignment, we conducted ablation studies by training and evaluating the model with or without aligning using \datasetnamefull~ respectively.
We conduct experiments on various multimodal models, including both multimodal language models and CLIP models: \modelnamefull~, InternVL2-8B \citep{chen2024far}, MiniCPM-V-2.6-8B \citep{yao2024minicpm}, and PubMedCLIP \citep{eslami2023pubmedclip}.
As shown in~\cref{tab:model_impro}, incorporating multigranular alignment significantly enhances performance across all tested multimodal models. 
Notably, \modelnamefull~exhibited improvements of 2.0\%, 5.0\%, and 1.6\% on VQA-RAD, SLAKE, and PathVQA, respectively, compared to its counterpart without alignment. These findings underscore the potential of \modelnamefull~as a foundational dataset capable of enhancing the medical performance of various multimodal models.

\label{sec:benchmark}
\begin{table}[t!]  
\centering  
\begin{subtable}{1.0\textwidth} 
\centering 
\resizebox{\linewidth}{!}{
\begin{tabular}{ll|cll|cll|cll}  
 & & \multicolumn{3}{c|}{\bf VQA-RAD} & \multicolumn{3}{c|}{\bf SLAKE} & \multicolumn{3}{c}{\bf PathVQA}  \\
Method   &  & Ref   & Open   & Closed   & Ref   & Open   & Closed    & Ref  & Open &  Closed  \\ \hline

  \multicolumn{11}{l}{\it Results of Non-MLLM Based Methods} \\  
\multicolumn{2}{l|}{VL Encoder--Decoder~\cite{bazi2023vision}} &  71.5 & & 82.5 &  & &  & 71.5 & & 85.6\\
\multicolumn{2}{l|}{Q2ATransformer~\cite{liu2023q2atransformer}} &  79.2 & & 81.2 & & &  & 54.9 & & 88.9 \\
\multicolumn{2}{l|}{Prefix T. Medical LM~\cite{van2023open}} &  & & & 84.3 & & 82.0 & 40.0 & & 87.0\\ 
\multicolumn{2}{l|}{PubMedCLIP~\cite{eslami2023pubmedclip}} &  60.1 & & 80.0 & 78.4 & & 82.5 & & & \\
\multicolumn{2}{l|}{BiomedCLIP~\cite{zhang2023large}} &  67.6 & & 79.8 & 82.1 & & 89.7 & & & \\
\multicolumn{2}{l|}{M2I2~\cite{li2022self}} & 66.5 & & 83.5 & 74.7 & & 91.1 & 36.3 & & 88.0 \\ \hline

\multicolumn{11}{l}{\it Results of MLLM Based Methods} \\
\multicolumn{2}{l|}{GPT-4V~\cite{achiam2023gpt}} &  &  39.5 & 78.9 &  & 33.6 & 43.6 &  & - & - \\
\multicolumn{2}{l|}{LLaVA} &   & 50.0 & 65.1 & & 78.2 & 63.2 & & 7.7 & 63.2  \\ 
\multicolumn{2}{l|}{\llavamed} &  & 61.5 & 84.2 &  & 83.1 & 85.3 & & 37.9 & 91.2 \\

\multicolumn{2}{l|}{\modelnamefull} &  & \textbf{65.1} & \textbf{84.9} &  & \textbf{86.4} & \textbf{89.2} & & \textbf{37.9} & \textbf{92.7} \\
\end{tabular}}  

\end{subtable}  
\caption{\textbf{Comparison with prior state-of-the-art supervised methods.} For open-ended questions, non-multimodal language models still
formulate the problem as classification among distinct answers in the training set, which may overestimate their
generalizability as these datasets are unusual in that the test answers are almost always present in training.}
\label{tab:model_perf}  
\end{table}

\begin{table}[t]
\centering
\caption{\textbf{Comparison of different models with or without alignment pretraining with \datasetnamefull}. The notation w/ and w/o indicate models with and without pretraining on \datasetnamefull, respectively.}
\begin{subtable}{1.0\textwidth} 
\resizebox{\linewidth}{!}{
\begin{tabular}{c|c|c|c|c|c|c|c|c|c|c}
\toprule
\multirow{2}{*}{\textbf{Model}} & \multirow{2}{*}{\textbf{Dataset Use}} & \multicolumn{3}{c|}{\textbf{VQA-RAD}} & \multicolumn{3}{c|}{\textbf{SLAKE}} & \multicolumn{3}{c}{\textbf{PathVQA}} \\
 &  & \textbf{open} & \textbf{close} & \textbf{average} & \textbf{open} & \textbf{close} & \textbf{average} & \textbf{open} & \textbf{close} & \textbf{average} \\
\midrule
\multirow{2}{*}{\textbf{\modelnamefull}} & w/o & 63.3 & 82.7 & 73.0 & 81.0 & 84.5 & 82.8 & 35.1 & 92.3 & 63.7 \\
 & w/ & \textbf{65.1} {\footnotesize\color[HTML]{009901} \cgap{+}({1.8})} & \textbf{84.9} {\footnotesize\color[HTML]{009901} \cgap{+}({2.2})} & \textbf{75.0} {\footnotesize\color[HTML]{009901} \cgap{+}({2.0})} & \textbf{86.4} {\footnotesize\color[HTML]{009901} \cgap{+}({5.4})} & \textbf{89.2} {\footnotesize\color[HTML]{009901} \cgap{+}({4.7})} & \textbf{87.8} {\footnotesize\color[HTML]{009901} \cgap{+}({5.0})} & \textbf{37.9} {\footnotesize\color[HTML]{009901} \cgap{+}({2.8})} & \textbf{92.7} {\footnotesize\color[HTML]{009901} \cgap{+}({0.4})} & \textbf{65.3} {\footnotesize\color[HTML]{009901} \cgap{+}({1.6})} \\
\midrule
\multirow{2}{*}{\textbf{MiniCPM-V-2.6-8B~\citep{yao2024minicpm}}} & w/o & 48.5 & 86.4 & 67.5 & 57.2 & 80.0 & 68.6 & 31.2 & 90.5 & 60.9 \\
 & w/ 
 & 50.5 {\footnotesize\color[HTML]{009901} \cgap{+}({2.0})} 
 & 87.6 {\footnotesize\color[HTML]{009901} \cgap{+}({1.2})}
 & 69.1 {\footnotesize\color[HTML]{009901} \cgap{+}({1.6})}
 & 65.3 {\footnotesize\color[HTML]{009901} \cgap{+}({8.1})}
 & 80.6 {\footnotesize\color[HTML]{009901} \cgap{+}({0.6})}
 & 73.0 {\footnotesize\color[HTML]{009901} \cgap{+}({4.4})}
 & 34.2 {\footnotesize\color[HTML]{009901} \cgap{+}({3.0})}
 & 94.8 {\footnotesize\color[HTML]{009901} \cgap{+}({4.3})}
 & 64.5 {\footnotesize\color[HTML]{009901} \cgap{+}({3.6})}\\
\midrule
\multirow{2}{*}{\textbf{InternVL2-8B~\citep{chen2024far}}} & w/o & 38.2 & 76.2 & 57.2 & 61.7 & 77.8 & 69.8 & 16.8 & 86.4 & 51.6 \\
 & w/ 
 & 40.7 {\footnotesize\color[HTML]{009901} \cgap{+}({2.5})}
 & 80.0 {\footnotesize\color[HTML]{009901} \cgap{+}({3.8})}
 & 60.4 {\footnotesize\color[HTML]{009901} \cgap{+}({3.2})}
 & 66.4 {\footnotesize\color[HTML]{009901} \cgap{+}({4.7})}
 & 78.8 {\footnotesize\color[HTML]{009901} \cgap{+}({1.0})}
 & 72.6 {\footnotesize\color[HTML]{009901} \cgap{+}({2.8})}
 & 23.6 {\footnotesize\color[HTML]{009901} \cgap{+}({6.8})}
 & 87.4 {\footnotesize\color[HTML]{009901} \cgap{+}({1.0})}
 & 55.5 {\footnotesize\color[HTML]{009901} \cgap{+}({3.9})}\\
 \midrule
\multirow{2}{*}{\textbf{PubMedCLIP~\citep{eslami2023pubmedclip}}} & w/o & 55.6 & 79.3 & 67.5 & - & - & - & - & - & - \\
 & w/ 
 & 60.6 {\footnotesize\color[HTML]{009901} \cgap{+}({5.0})}
 & 79.7 {\footnotesize\color[HTML]{009901} \cgap{+}({0.4})}
 & 70.2 {\footnotesize\color[HTML]{009901} \cgap{+}({2.7})}
 & -
 & -
 & -
 & -
 & -
 & -\\
\bottomrule
\end{tabular}}
\end{subtable}
\label{tab:model_impro} 
\vspace{-0.6cm}
\end{table}

\section{Conclusion}\label{sec:conclusion}
\vspace{-3mm}
This paper introduces MedTrinity-25M, a large-scale multimodal medical dataset comprising over 25 million image-ROI-description triplets sourced from more than 30 online resources, spanning 10 modalities and covering over 65 diseases. We have developed the first automated pipeline to scale up multimodal data by generating multigranular visual and textual annotations from unpaired images. We believe that MedTrinity-25M's enriched annotations have the potential to support a wide range of multimodal tasks, such as captioning, report generation, classification, and segmentation, as well as facilitate the large-scale pre-training of multimodal medical AI models.
\section*{Acknowledgement}
We thank the Microsoft Accelerate Foundation Models Research Program, the OpenAI Researcher Access Program, TPU Research Cloud (TRC) program, Google Cloud Research Credits program, AWS Cloud Credit for Research program, and Lambda Cloud for supporting our computing needs.

\bibliographystyle{iclr2025_conference}
\bibliography{iclr2025_conference}

\clearpage
\appendix
\section*{Appendix}
\appendix
We present the following items in the Appendix:
\begin{enumerate}
    \item Data source about \datasetnamefull. (Section \ref{sec:datasource})

    \item Quantitative comparison between GPT-4V and \captioner~(Section \ref{sec:compare_gpt}).
   \item Examples of ROI for normal regions and multiple regions.(Section \ref{sec:example_roi}).
    \item The list of expert ROI models (Section \ref{sec:model_roi}).
    \item Details of LLM Evaluation of Alignment (Section \ref{sec:prompt_alignment}).
    \item Prompt for generating \datasetnamefull. (Section \ref{sec:prompt}).
\end{enumerate}

\section{Data Source}
\label{sec:datasource}

\setlength{\tabcolsep}{0.5pt}
\small

\begin{longtable}{p{6cm}lll|c|c|c|c}
\caption{Data sources for \datasetnamefull~from various medical image datasets, detailing their modalities, biological structures, quantities, and annotations.}
 \\
 
  \hline
    \textbf{Dataset Name} & \textbf{Modality} & \textbf{\begin{tabular}[c]{@{}l@{}}Biological\\Structures\end{tabular}} & \textbf{Quantity} & \textbf{Text} & \textbf{\begin{tabular}[c]{@{}l@{}}Disease\\Type\end{tabular}} & \textbf{BBox} & \textbf{Mask} \\ \hline\hline
    \endfirsthead
    \multicolumn{8}{c}{{\tablename\ \thetable{} : Continued from previous page}} \\ \hline\hline
    \textbf{Dataset Name} & \textbf{Modality} & \textbf{\begin{tabular}[c]{@{}l@{}}Biological\\Structures\end{tabular}} & \textbf{Quantity} & \textbf{Text} & \textbf{\begin{tabular}[c]{@{}l@{}}Disease\\Type\end{tabular}} & \textbf{BBox} & \textbf{Mask} \\ \hline\hline
    \endhead
    BHX\citep{bhx} & MRI & brain & 973908 & \ding{55} & \ding{55} & \ding{55} & \checkmark \\ \hline
    BRATS24-MICCAI\citep{de2024bratsmiccai} & MRI & brain & 2535132 & \ding{55} & \ding{55} & \checkmark & \ding{55} \\ \hline
    BRATS-ISBI\citep{Karargyris2023-tu-brats24-1-isbi} & MRI & brain & 987340 & \ding{55} & \ding{55} & \checkmark & \ding{55} \\ \hline
    breast histopathology\citep{Janowczyk2016-ap-breasthistopathology,Cruz-Roa2014-xg-breasthistopathology-2} & Histopathology & breast & 547403 & \ding{55} & \checkmark & \ding{55} & \ding{55} \\ \hline
    BreastCancer\citep{Ding2022-iw-Breastcancer} & Histopathology & breast & 1824 & \ding{55} & \ding{55} & \checkmark & \ding{55} \\ \hline
    CheXpert\citep{irvin2019chexpert} & X-Ray & lung & 183242 & \ding{55} & \checkmark & \ding{55} & \ding{55} \\ \hline
    CISC\citep{gamper2020pannukecisc} & Histopathology & \begin{tabular}[c]{@{}l@{}}Adrenal,\\ Bile duct,\\ Bladder,\\ Breast,\\ Colon, \\Cervix,\\ Esophagus\\ Kidney,\\ Liver,etc \end{tabular} & 16285 & \ding{55} & \checkmark & \checkmark & \ding{55} \\ \hline
    CPD\citep{Wagner2022-bz-cpd} & Histopathology & skin & 204 & \ding{55} & \ding{55} & \checkmark & \ding{55} \\ \hline
    CT-RATE\citep{hamamci2024foundation-ctrate} & CT & \begin{tabular}[c]{@{}l@{}}lung,\\ liver,\\ mediastinum, \\kidney, \\heart, etc.\end{tabular} & 3869640 & \checkmark & \ding{55} & \ding{55} & \ding{55} \\ \hline
    DeepLesion\citep{yan2017deeplesion} & CT & \begin{tabular}[c]{@{}l@{}}bone,\\ abdomen,\\ mediastinum,\\ liver,\\ lung,\\ kidney,\\ soft tissue, \\pelvis\end{tabular} & 2889672 & \ding{55} & \ding{55} & \ding{55} & \checkmark \\ \hline
    FLARE23\citep{Ma_Wang_2023_flare} & CT & \begin{tabular}[c]{@{}l@{}}Liver,\\ kidney, \\spleen,\\ pancreas,\\ Aorta, \\adrenal gland, \\Gallbladder, \\esophagus,\\ stomach,\\ duodenum,etc.\end{tabular} & 13770 & \ding{55} & \checkmark & \checkmark & \ding{55} \\ \hline
    ihc4bc\citep{Akbarnejad2023-ry-ihc4bc} & Microscopy & cell & 102535 & \ding{55} & \checkmark & \ding{55} & \ding{55} \\ \hline
    KIPA22\citep{Shao2012-nc-kipa1,Shao2011-va-kipa2,He2020-qj-kipa3,He2021-dk-kipa4} & CT & \begin{tabular}[c]{@{}l@{}} kidney,\\ cervix \end{tabular}& 26878 & \ding{55} & \ding{55} & \checkmark & \ding{55} \\ \hline
    LLaVA-Med\citep{li2024llavamed} & \begin{tabular}[c]{@{}l@{}}CT,\\ MR,\\ Endoscopy,\\ X-Ray, \\Ultrasound,\\ Histopathology, \\Dermoscopy, \\Microscopy, \\Fundus,\\ PET\end{tabular} & \begin{tabular}[c]{@{}l@{}}cell, rib,\\ tissue, \\face,\\ brain, \\vascular,\\ liver,\\ bone,\\ lymph, etc.\end{tabular} & 22550 & \checkmark & \ding{55} & \ding{55} & \ding{55} \\ \hline
    LLD-MMRI\citep{lou2024sdrformer-lld-mmri} & MRI & liver & 21523 & \ding{55} & \ding{55} & \checkmark & \ding{55} \\ \hline
    MAMA-MIA\citep{garrucho2024mama-mia} & MRI & breast & 316113 & \ding{55} & \ding{55} & \checkmark & \ding{55} \\ \hline
    MIMIC-CXR-JPG\citep{johnson2019mimiccxrjpg} & X-Ray & lung& 240506 & \checkmark & \checkmark & \ding{55} & \checkmark \\ \hline
    NCT-CRC-HE-100K\citep{Kather2018-aj-crc100k} & Histopathology & colon & 100361 & \ding{55} & \checkmark & \ding{55} & \ding{55} \\ \hline
    NIH-CXR\citep{Wang2017-lg-nihcxr3,Wang2017-ra-nihcxr1,Wang2019-gu-nihcxr2} & X-Ray & lung & 986 & \ding{55} & \ding{55} & \ding{55} & \checkmark \\ \hline
    PadChest\citep{Bustos_2020-padchest} & CT & lung & 96284 & \checkmark & \ding{55} & \ding{55} & \ding{55} \\ \hline
    PatchGastricADC22\citep{Tsuneki2021-ft-patchgastricadc22} & MRI & brain & 98399 & \ding{55} & \checkmark & \ding{55} & \ding{55} \\ \hline
    Path-VQA training\citep{he2020pathvqa} & Pathology & \begin{tabular}[c]{@{}l@{}}gastrointestinal,\\colon,\\appendix,\\pinworm,etc. \end{tabular}& 13375 & \checkmark & \checkmark & \ding{55} & \ding{55} \\ \hline
    PMC-OA\citep{noauthor_undated-bj-pmc-oa} & \begin{tabular}[c]{@{}l@{}}CT,\\ MR, \\Endoscopy,\\ X-Ray, \\Ultrasound, \\Histopathology, \\Dermoscopy,\\ Microscopy, \\Fundus, \\PET\end{tabular} & \begin{tabular}[c]{@{}l@{}}cell,\\ tissue,\\ vascular,\\ brain, \\bone,\\ liver,\\ lymph,\\ eye, \\epithelium, etc.\end{tabular} & 856999 & \checkmark & \ding{55} & \ding{55} & \ding{55} \\ \hline
    PMC-VQA\citep{zhang2023pmcvqa} & \begin{tabular}[c]{@{}l@{}}CT,\\ MR,\\ Endoscopy,\\ X-Ray, \\Ultrasound, \\Histopathology, \\Dermoscopy,\\ Microscopy,\\ Fundus, \\PET\end{tabular} & \begin{tabular}[c]{@{}l@{}}cell,\\ brain,\\ tissue,\\ artery,\\ bone,\\ face,\\ rib, \\vascular,\\liver,\\ eye, etc.\end{tabular} & 144999 & \checkmark & \ding{55} & \ding{55} & \ding{55} \\ \hline
    PTCGA\citep{kawai2023largeptcga} & Histopathology & \begin{tabular}[c]{@{}l@{}}brain,\\ breast, \\uterine corpus,\\ kidney,\\ lung,\\ thyroid\end{tabular} & 3293965 & \ding{55} & \checkmark & \checkmark & \ding{55} \\ \hline
    Quilt-1M\citep{quilt} & Histopathology & \begin{tabular}[c]{@{}l@{}}skin, \\lung,\\ soft tissue,\\ blood,\\ kidney,\\ bone, etc.\end{tabular} & 643819 & \checkmark & \ding{55} & \ding{55} & \ding{55} \\ \hline
    SAMMed-20M\citep{sam20m} & \begin{tabular}[c]{@{}l@{}}X-Ray,\\ PET,\\ CT,\\ MR,\\ Endoscopy, \\dermoscopy\end{tabular} & \begin{tabular}[c]{@{}l@{}}brain,\\ kidney,\\ liver,\\ lung, \\pancreas,\\ pulmonary,\\ hepatic, \\skin, etc.\end{tabular} & 5491274 & \ding{55} & \checkmark & \checkmark & \ding{55} \\ \hline
    SLAKE training\citep{liu2021slake} & \begin{tabular}[c]{@{}l@{}}CT,\\ MRI,\\ X-Ray\end{tabular} & \begin{tabular}[c]{@{}l@{}}brain,\\ liver,\\ kidney,\\pelvic,\\ lung\end{tabular} & 646 & \checkmark & \checkmark & \checkmark & \ding{55} \\ \hline
    TCGA\citep{kawai2023largeptcga} & Histopathology&tissue & 1142221 & \ding{55} & \ding{55} & \checkmark & \ding{55} \\ \hline
    ULS23 & CT & \begin{tabular}[c]{@{}l@{}}lung, \\lymph nodes,\\ bladder,\\ brain, \\colon,\\  kidney, \\lung. \\\end{tabular} & 105669 & \ding{55} & \ding{55} & \checkmark & \ding{55} \\ \hline
    
    VALSET\citep{tolkach2023artificialvalset} & Histopathology &\begin{tabular}[c]{@{}l@{}} oesophagus,\\stomach \end{tabular}& 277565 & \ding{55} & \checkmark & \ding{55} & \ding{55} \\ \hline
    VQA-RAD training\citep{lau2018datasetvqarad} & \begin{tabular}[c]{@{}l@{}}X-Ray,\\ MRI\end{tabular} &\begin{tabular}[c]{@{}l@{}}brain,\\lung,\\abdomen,etc.\end{tabular}& 1758 & \checkmark & \checkmark & \ding{55} & \ding{55} \\ \hline\hline
    &&&&&&&\\
    \textbf{Total} &&& \textbf{25016845} & &&& \\ \hline\hline
    \hline
    \label{tab:all_datasets}
\end{longtable}

\section{Quantitative Comparison of \captioner~with GPT-4V}
\label{sec:compare_gpt}
As detailed in Section 3.2.2 of the main paper, we developed an enhanced version of LLaVA~\citep{li2024llava}, called \captioner. This enhancement leverages the latest \llama~\citep{llama3} to boost linguistic capabilities and incorporates multi-scale feature extraction \citep{shi2024we} to improve vision capabilities.

\begin{figure*}[t]
  \centering
  \includegraphics[page=1, width=\textwidth]{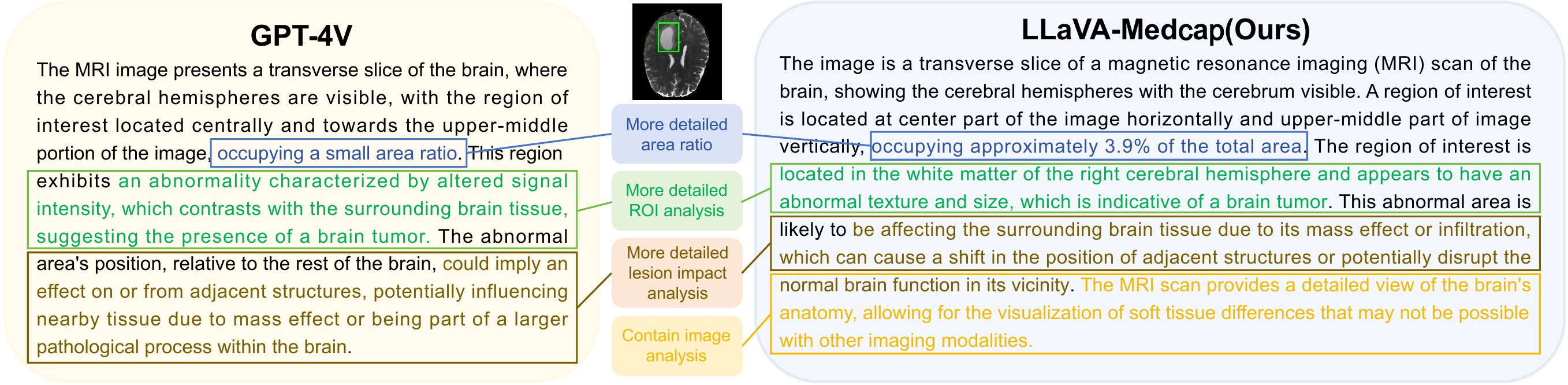}
  \caption{
  \textbf{Qualitative Comparison with sample generated by GPT-4V.} Compared to GPT-4V, our model generate more detailed caption.
  }
  \label{fig:compare_gpt}
\end{figure*}

To justify the selection of our specialized medical model, \modelnamefull, over GPT-4V for generating textual descriptions, we conducted a quantitative comparison of the outputs generated by both models. We assessed the level of detail by comparing the average word count of text descriptions generated for the same sample. \modelnamefull, after task-specific fine-tuning, outperformed GPT-4V by 3.6\% in word count, indicating that the descriptions generated by \captioner~are more detailed. We also provide a qualitative comparison with a sample generated by \modelnamefull~and GPT-4V in \cref{fig:compare_gpt}.
Based on these findings, we selected \captioner~to generate \datanamefull~ for our entire \datasetnamefull.

\section{Examples of ROIs}
\label{sec:example_roi}
\begin{figure*}[t]
  \centering
  \begin{subfigure}{0.40\textwidth}
    \centering
        
    \includegraphics[page=1, width=0.6\textwidth]{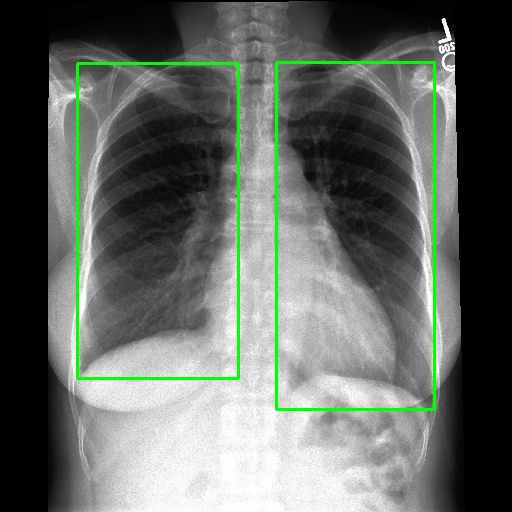}
    \label{fig:normal_roi_1}
    \caption{A no infection sample from MIMIC-CXR. The ROIs highlight the left and right lungs.}
  \end{subfigure}
  \hfill
  \begin{subfigure}{0.40\textwidth}
    \centering
        
    \includegraphics[page=2, width=0.6\textwidth]{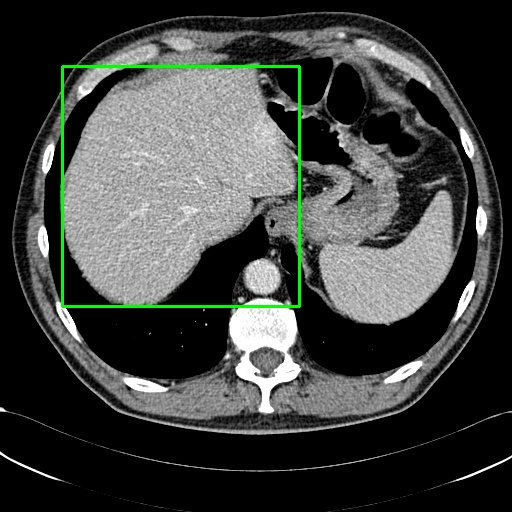}
\caption{A healthy sample from SLAKE. The ROI points out the liver.}
    \label{fig:normal_roi_2}
  \end{subfigure}
  \caption{Examples of ROIs for normal regions.}
  \label{fig:normal_roi_combined}
\end{figure*}

As described in Section 3.1 of the main paper, the ROIs identified by expert grounding models predominantly capture pathological features such as lesions, inflammation, neoplasms, infections, or other potential abnormalities. In rare cases where no abnormalities are found, the ROIs typically focus on the primary object or organ in the image. Examples of such normal ROIs are presented in~\cref{fig:normal_roi_combined}.

\begin{figure*}[t]
  \centering
  \includegraphics[width=\textwidth]{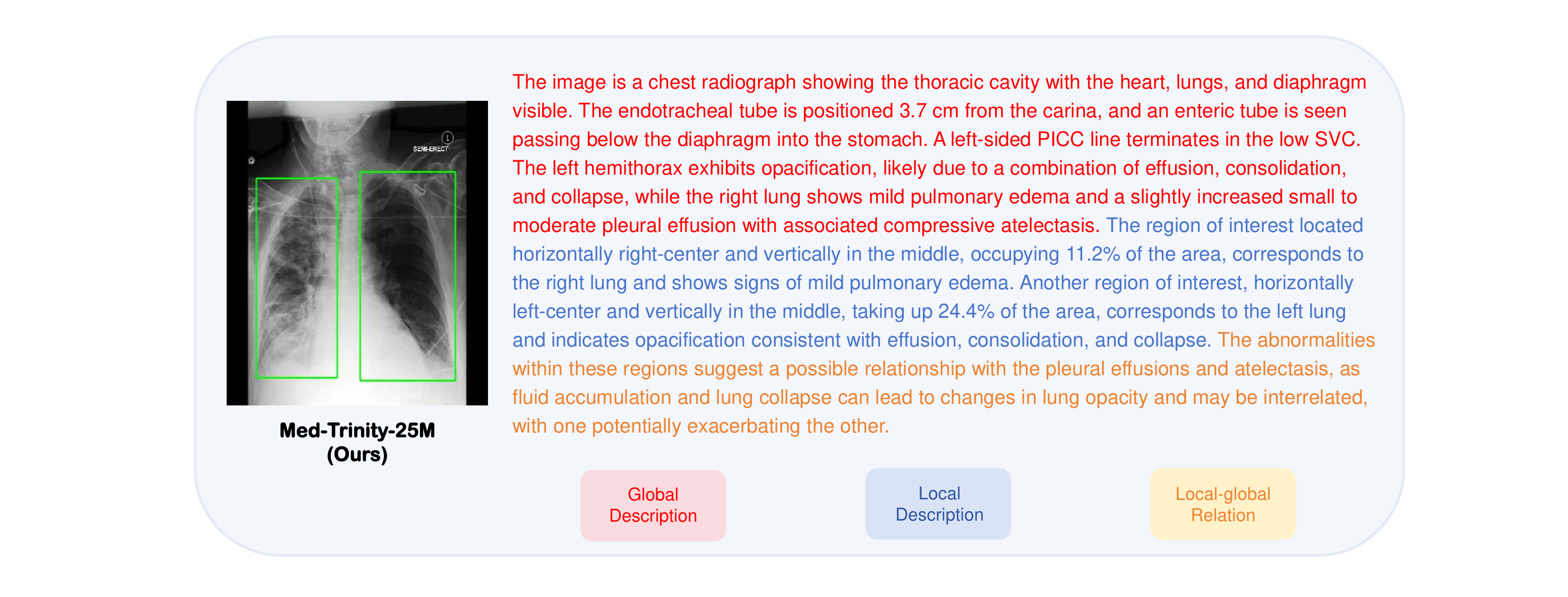}

  \caption{\textbf{A chest radiography example where global information matters.} The diagnosis in this case requires a comprehensive analysis of the entire image, encompassing both the left and right lungs. Here, ROIs encompass the large lesion areas of left and right lungs. Detailed local texture analysis of each region contributes to the overall global diagnosis}  
    \label{fig:global_local_example}
\end{figure*}
In some instances where global context is also critical for disease identification, the ROIs encompass multiple lesion areas, integrating both global and local information. For example, in chest radiography, analyzing both lungs and their overall structure is often essential for accurate diagnosis, as shown in~\cref{fig:global_local_example}. By providing multigranular annotations that incorporate both local and global perspectives, our dataset helps multigranular alignment for medical foundation models.

\section{List of Expert models to locate ROIs}
\label{sec:model_roi}

\setlength{\tabcolsep}{12pt}
\begin{table}[t]
  \centering
  \caption{List of expert models used to generate ROIs for different datasets.}
  \begin{tabular}{c|l|c}
    \toprule
    ID & Dataset Name & Model \\
    \midrule
    1 & breast histopathology & \multirow{10}{*}{HoverNet~\citep{stringer2024cellpose3}} \\
    2 & BreastCancer &  \\
    3 & CISC &  \\
    4 & CPD &  \\
    5 & NCT-CRC-HE-100K &  \\
    6 & PTCGA &  \\
    7 & TCGA &  \\
    8 & VALSET &  \\
    9 & ihc4bc &  \\
    10 & Quilt-1M &  \\
    \midrule
    11 & CT-RATE & \multirow{1}{*}{SAT~\citep{zhao2023one}} \\
    \midrule
    12 & PMC-OA & \multirow{4}{*}{DINO~\citep{caron2021emerging}} \\
    13 & PMC-VQA &  \\
    14 & LLaVA-Med &  \\
    15 & Path-VQA training &  \\
    \midrule
    16 & PadChest & \multirow{3}{*}{CheXmask~\citep{gaggion2023chexmask1PhysioNet}~\citep{Gaggion_2022_chexmask2}} \\
    17 & MIMIC-CXR-JPG &  \\
    18 & CheXpert & \\
    \bottomrule
  \end{tabular}%
  \label{tab:model_roi}%
\end{table}%

As detailed in Section 3.2.1 of the main paper, for datasets lacking localization information such as segmentation masks and bounding boxes, we employ various pretrained expert models to identify the ROIs. The specific expert models used for each dataset are listed in \cref{tab:model_roi}.

\section{Details of LLM Evaluation of Alignment}

\begin{figure*}[t]
  \centering
  
  \includegraphics[page=1, width=\textwidth]{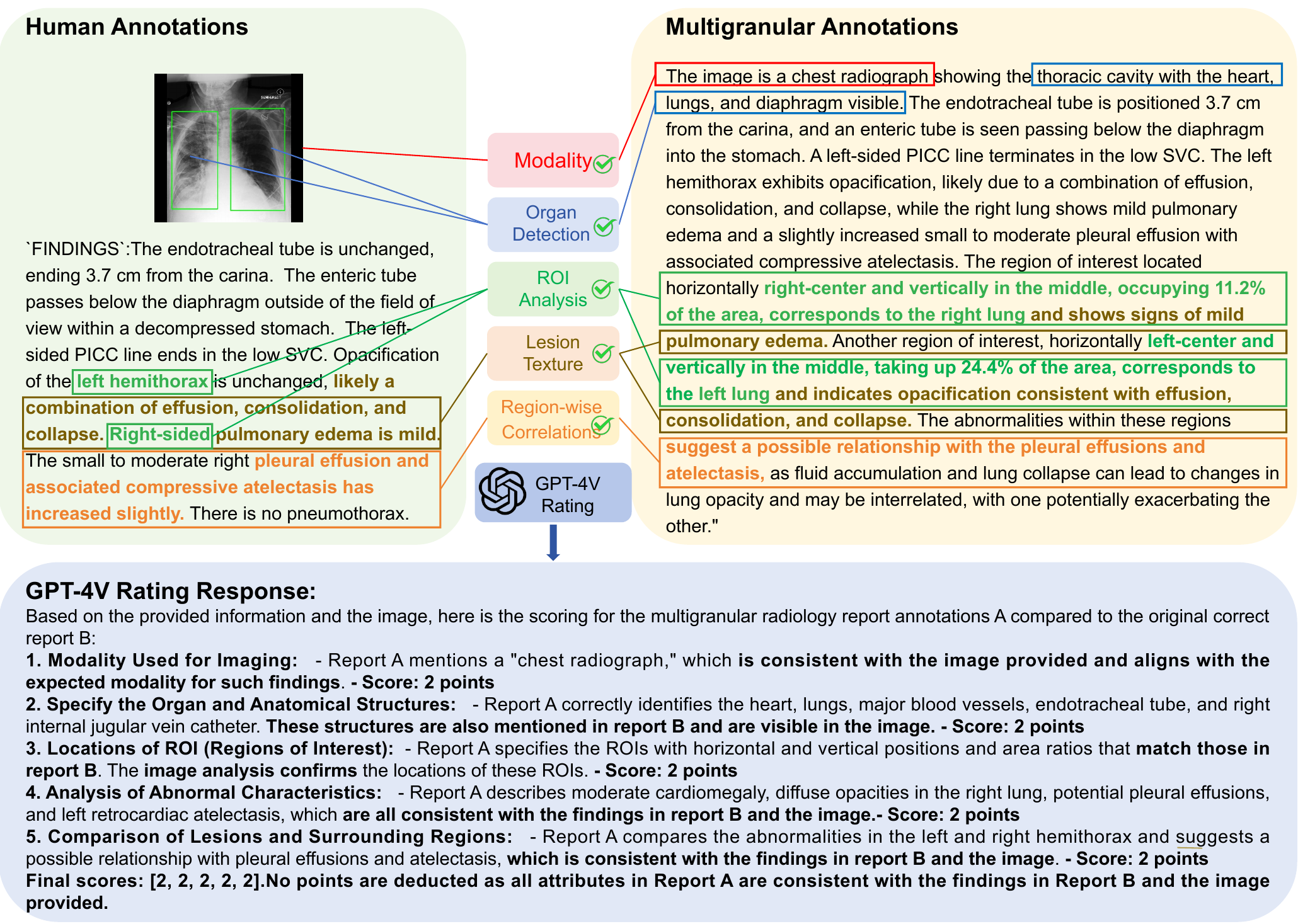}
\caption{\textbf{An example of a perfect score result evaluated by GPT-4V.} GPT-4V assesses five criteria, each fully aligned with human annotations, resulting in perfect scores.}  
  \label{fig:alignment_example}
\end{figure*}
An example of perfect alignment score results evaluated by GPT-4V is shown in \cref{fig:alignment_example}. In these examples, GPT-4V fully aligned with human annotations across all five criteria, resulting in perfect alignment scores. The prompt used to query GPT-4V for evaluating the alignment score is shown in \cref{fig:alignment_prompt} of supplementary.

\label{sec:prompt_alignment}
The prompt used to query GPT-4V for evaluating the alignment score is shown in \cref{fig:alignment_prompt}.
\begin{figure*}[htbp]
  \centering

  \includegraphics[page=1, width=0.7\textwidth]{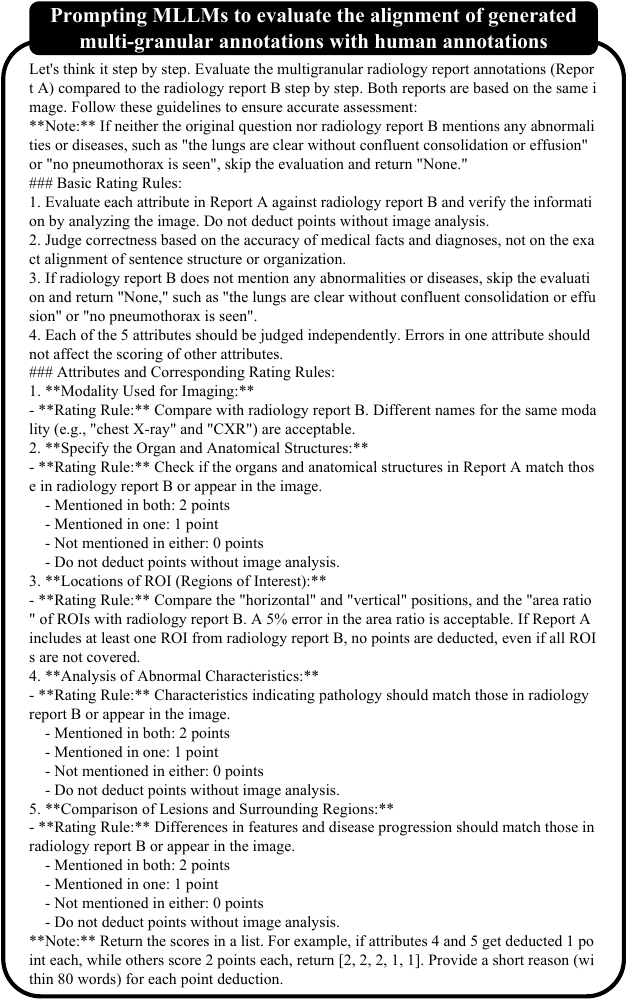}

    \caption{Prompt used to evaluate the alignment of generated \datanamefull.}
      \label{fig:alignment_prompt}
\end{figure*}

\section{Prompt Template for Generation of Multigranular Text Description}
\label{sec:prompt}
\begin{figure*}[htbp]
  \centering
  \includegraphics[page=1, width=0.68\textwidth]{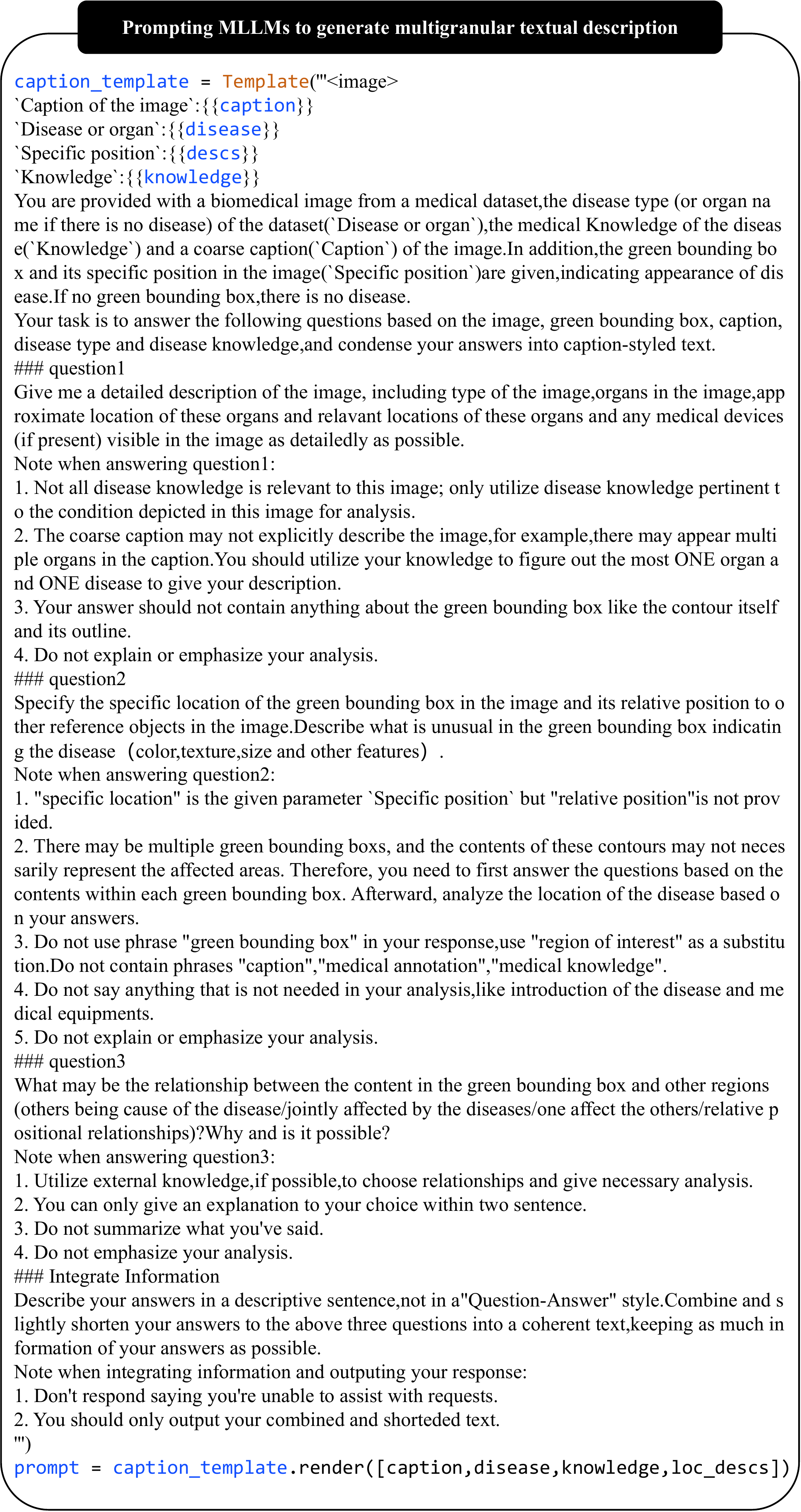}

  \caption{Prompt used to generate multigranular annotations of \datanamefull.}
  \label{fig:prompt}
\end{figure*}
To generate multigranular textual descriptions, we design a multi-task prompting approach, breaking down this task into several smaller descriptive tasks. The model's responses to these different tasks collectively form the final fine-grained text description.

\cref{fig:prompt} illustrates our prompt template consisting of a three-level hierarchical framework with questions to instruct \llmname:

\textbf{Step 1 - Global Understanding}: Instruct \llmname~to provide a comprehensive description of the image, detailing all modalities, identified anatomical structures, and their approximate locations. This step ensures that \llmname~gains an overarching understanding and basic information about the image.

\textbf{Step 2 - Local Analysis}: Instruct \llmname~to conduct a detailed analysis of the regions of interest (ROI), including their locations, abnormalities, and textures. This step guides \llmname~to focus on specific lesions for a thorough assessment.

\textbf{Step 3 - Region-wise Correlations}: Instruct \llmname~to examine the relationship between different regions and predict how the surrounding areas will be affected by the lesions in the ROI. This step aims to understand the interaction between local and global attributes, assessing the impact of local abnormalities on the entire organ for accurate disease diagnosis.

\clearpage

\label{sec:appendix}

\clearpage

\end{document}